\documentclass[letterpaper]{article} 
\usepackage{aaai2026}
\usepackage{times}  
\usepackage{helvet}  
\usepackage{courier}  
\usepackage[hyphens]{url}  
\usepackage{graphicx} 
\usepackage{natbib}  
\usepackage{caption} 
\PassOptionsToPackage{table}{xcolor}
\usepackage[utf8]{inputenc} 
\usepackage[T1]{fontenc}    
\usepackage{booktabs}       
\usepackage{amsfonts}       
\usepackage{nicefrac}       
\usepackage{microtype}      
\usepackage{float}
\usepackage{epsfig}
\usepackage{graphicx}
\usepackage{amsmath}
\usepackage{amssymb}
\usepackage{multirow}
\usepackage{array}
\usepackage{tikz}
\usepackage{pgfplots}
\pgfplotsset{compat=1.18}
\usepackage{subcaption}
\usepackage{mathrsfs}
\usepackage{enumitem}
\newtheorem{theorem}{Theorem}

\usepackage{newfloat}
\usepackage{listings}
\DeclareCaptionStyle{ruled}{labelfont=normalfont,labelsep=colon,strut=off} 
\floatstyle{ruled}
\newfloat{listing}{tb}{lst}{}
\floatname{listing}{Listing}
\title{CCD: Continual Consistency Diffusion for Lifelong Generative Modeling}
\author{
    Jingren Liu\textsuperscript{\rm 1,4},
    Shuning Xu\textsuperscript{\rm 2},
    Yun Wang\textsuperscript{\rm 3,4},
    Zhong Ji\textsuperscript{\rm 1},
    Xiangyu Chen\textsuperscript{\rm 4}
}
\affiliations{
    \textsuperscript{\rm 1}Tianjin University\\
    \textsuperscript{\rm 2}University of Macau\\
    \textsuperscript{\rm 3}City University of Hong Kong\\
    \textsuperscript{\rm 4}Institute of Artificial Intelligence (TeleAI), China Telecom\\
    jrl0219@tju.edu.cn, 
    yc07425@um.edu.mo, 
    ywang3875-c@my.cityu.edu.hk,\\
    jizhong@tju.edu.cn,
    chxy95@gmail.com
}

\usepackage{bibentry}

\begin{document}

\maketitle

\begin{abstract}
While diffusion-based models have shown remarkable generative capabilities in static settings, their extension to continual learning (CL) scenarios remains fundamentally constrained by Generative Catastrophic Forgetting (GCF). We observe that even with a rehearsal buffer, new generative skills often overwrite previous ones, degrading performance on earlier tasks. Although some initial efforts have explored this space, most rely on heuristics borrowed from continual classification methods or use trained diffusion models as ad hoc replay generators, lacking a principled, unified solution  to mitigating GCF and often conducting experiments under fragmented and inconsistent settings. To address this gap, we introduce the Continual Diffusion Generation (CDG), a structured pipeline that redefines how diffusion models are implemented under CL and enables systematic evaluation of GCF. Beyond the empirical pipeline, we propose the first theoretical foundation for CDG, grounded in a cross-task analysis of diffusion-specific generative dynamics. Our theoretical investigation identifies three fundamental consistency principles essential for preserving knowledge in the rehearsal buffer over time: inter-task knowledge consistency, unconditional knowledge consistency, and prior knowledge consistency. These criteria expose the latent mechanisms through which generative forgetting manifests across sequential tasks. Motivated by these insights, we further propose \textit{Continual Consistency Diffusion} (CCD), a principled training framework that enforces these consistency objectives via hierarchical loss functions: $\mathcal{L}_{IKC}$, $\mathcal{L}_{UKC}$, and $\mathcal{L}_{PKC}$. This framework fosters long-term retention of generative knowledge and stable integration of new capabilities. Extensive experiments show that CCD achieves state-of-the-art performance across various benchmarks, especially improving generative metrics in overlapping-task scenarios.
\end{abstract}

\section{Introduction}
\label{sec:intro}
\begin{figure}[h]
    \centering
    \includegraphics[scale=1.0]{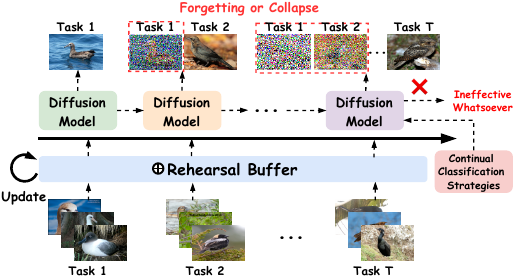}
    \caption{An overview of CDG pipeline and its challenges, highlighting the ineffectiveness of current continual classification strategies in preventing performance degradation and GCF in streaming tasks with diffusion models.}
    \label{Fig1}
    \vspace{-3ex}
\end{figure}

The remarkable success of diffusion models in synthesizing high-fidelity text and images \cite{bruce2024genie,nie2025large,yang2025mmada} has significantly accelerated the arrival of generative artificial intelligence (AGI). However, the static nature of their training pipeline hinders further advancement in dynamic real-world scenarios, such as personalized content creation or real-time virtual environment generation for interactive applications. When faced with the arrival of new data, existing approaches generally involve retraining models on both historical and current data to prevent significant performance degradation or generative collapse \footnote{During the training of streaming tasks, we observe that diffusion models occasionally experience sudden catastrophic failures on certain tasks, leading to a significant degradation.} on previously learned tasks. However, this is computationally expensive and results in considerable waste of both computational resources and energy, exacerbating the practical challenges of deploying diffusion models in continual learning (CL) contexts. These challenges are further compounded by the fact that current research on diffusion generation in CL remains fragmented, lacking a standardized experimental protocols \cite{zajkac2023exploring, masip2023continual, cheng2024semi} and thorough understanding. This underscores the urgent need for a systematic pipeline capable of quantifying the corresponding continual generative performance in a standardized manner. Accordingly, we present a formally grounded Continual Diffusion Generation (CDG) pipeline (see Figure.~\ref{Fig1}), upon which our theoretical analysis and experimental validation are based.

Moreover, despite recent efforts proposing various strategies to mitigate the generative catastrophic forgetting (GCF) in generative models, most remain misaligned with the core dynamics of diffusion-based architectures. Specifically, many methods rely on heuristics adapted from continual classification methods \cite{sun2024create, pfulb2021continual, varshney2021cam, zhang2024clog, masip2023continual, zajkac2023exploring, cheng2024semi}, employ stability-plasticity trade-offs originally designed for Generative Adversarial Networks (GANs) \cite{ali2025cfts, ye2021lifelong, ye2021lifelong2, gu2020association, zhao2024invertavatar}, or use trained diffusion models as ad hoc replay generators \cite{gao2023ddgr}. However, these techniques often conflict with the stochastic differential equations (SDEs) that govern the generative processes of diffusion models, leading to noticeable degradation in output quality. Empirical results confirm these limitations, as reflected in consistent declines in generative fidelity across sequential tasks, measured by the metric like Fréchet Inception Distance (FID) \cite{heusel2017gans} (see Figure.~\ref{Fig1} and Table.~\ref{tab:results}). In response to these research limitations, we begin by analyzing the unique mechanisms of diffusion models and investigate how they can be optimized in the context of streaming tasks, aiming to preserve shared knowledge across tasks and mitigate GCF.

To achieve this goal, we begin by formalizing diffusion trajectories, complex sequences of denoising operations that progressively transform noise into structured outputs, and examining how they interact through shared knowledge under streaming task scenarios. Grounded in Bayesian theory and multi-task learning principles \cite{yu2020gradient}, we derive a theoretical upper bound that quantifies the retention of generative knowledge across tasks in CDG, as formally defined in Theorem~\ref{thm:cross-task-diffusion}. Attaining this upper bound requires satisfying three critical consistency constraints: inter-task alignment of model-internalized knowledge, consistency in the mean embeddings of unconditional generated samples across tasks, and semantic consistency within the prior (i.e., label) space of original samples \footnote{Notably, the label space acts as a proxy for human prior knowledge, with the constraint aimed at preserving shared semantic structure across tasks, rather than improving classification performance.} across tasks.

Operationalizing the theoretical insights, we introduce the \textit{Continual Consistency Diffusion} (CCD) framework, which translates the derived guarantees into a tractable, consistency-driven optimization strategy. CCD enforces cross-task stability through a hierarchical integration of consistency objectives. Specifically, \emph{Inter-task Knowledge Consistency} regularizes model-internal knowledge representations across tasks, serving as the foundation for long-term retention. \emph{Unconditional Knowledge Consistency} preserves intrinsic generative behavior in the absence of explicit human priors, ensuring temporal coherence in the denoising process. Complementarily, \emph{Prior Knowledge Consistency} promotes alignment in the prior space by enforcing semantic correspondence between original samples across tasks. Collectively, these mechanisms move beyond standard regularization or classification heuristics. By directly constraining the geometric structure of the diffusion process, CCD enables robust and theoretically grounded continual generation in the CDG pipeline.

In summary, our work bridges the critical gap between traditional static diffusion models and the dynamic nature of real-world data streams. We present three main contributions. First, to the best of our knowledge, we are the first to rigorously formulate the CDG pipeline from a diffusion perspective.
Second, we establish the first theoretical framework for CDG rooted in SDE calculus, introducing novel stability bounds for SDE trajectories under sequential task adaptation (Theorem~\ref{thm:cross-task-diffusion}). Third, guided by these insights, we propose the CCD framework, which enforces intrinsic knowledge consistency through three synergistic components. Experiments on various benchmarks demonstrate its superiority, yielding significant gains while effectively mitigating GCF problem.

\section{Related Works}
\label{sec:related_works}
\noindent\textbf{Diffusion Models.}
Diffusion-based generative architectures have redefined state-of-the-art performance in structured data synthesis, primarily due to their ability to invert stochastic denoising trajectories. At a foundational level, these models learn to reverse-engineer discrete Markov chains (DDPMs) \cite{ho2020denoising} or continuous-time stochastic differential equations (SDEs) \cite{song2021scorebased}. Recent advancements in noise scheduling \cite{nichol2021improved, lu2022dpm} and adaptive sampling techniques \cite{lu2022dpm, zheng2023dpm} have further enhanced output fidelity. Such developments underscore diffusion models' theoretical strength as universal data approximators \cite{song2020denoising}, regardless of their discrete or continuous formulation \cite{bruce2024genie,nie2025large,yang2025mmada}. However, their success heavily relies on closed-world assumptions, where training data remains static and entirely observable. Consequently, a critical challenge persists in adapting diffusion models to dynamic, streaming data environments, paralleling incremental human cognition.

\noindent\textbf{Continual Classification.}  
The continual classification (CC) aims to enable models to progressively acquire new classification knowledge while retaining previously learned information, addressing the challenge of Catastrophic Forgetting (CF) \cite{kirkpatrick2017overcoming, li2017learning, parisi2019continual}. Traditional CC methods include replay-based techniques, which store subsets of historical data to maintain stable performance \cite{rolnick2019experience}, and regularization methods like EWC \cite{kirkpatrick2017overcoming} and LwF \cite{li2017learning}, which impose constraints on parameters to reduce interference between tasks. Additionally, gradient-based strategies such as GEM \cite{lopez2017gradient} orthogonalize gradients to minimize task conflicts. Recent advancements have focused on using generators trained on prior tasks as buffers, with DDGR \cite{gao2023ddgr} as a prominent example. While effective in mitigating CF in CC tasks, DDGR incurs substantial training overhead due to the need to synthesize past samples during each training batch.

In parallel, the advent of pre-training has driven progress in parameter-efficient fine-tuning techniques. Methods such as L2P \cite{wang2022learning} and DualPrompt \cite{wang2022dualprompt} utilize task-specific prompts to effectively balance adaptability and knowledge retention. Techniques like S-Prompt \cite{wang2022s} and CODA-Prompt \cite{smith2023coda} enhance performance by explicitly capturing domain relationships, while dynamic methods like DAP \cite{jung2023generating} and hierarchical approaches like HiDe-Prompt \cite{wang2024hierarchical} support adaptation across diverse domains. 
Despite these advancements, the existing CC research remains focused on basic classification tasks, limiting applicability to complex real-world scenarios. To bridge this gap, we investigate extending CC methods to practical and challenging applications, namely continual generation, within our standardized CDG pipeline.

\noindent\textbf{Continual Generation.} The continual generation represents a significant blind spot in the current landscape of CL research, with only a scant body of work dedicated to this area. Among these studies, most have primarily explored methods based on GAN architectures \cite{ali2025cfts,ye2021lifelong,ye2021lifelong2,gu2020association,zhao2024invertavatar}. Additionally, many approaches attempt to adapt techniques originally developed for CC tasks \cite{sun2024create,pfulb2021continual,varshney2021cam,zhang2024clog,masip2023continual}, applying relevant fine-tuning strategies. However, given that generation and classification are fundamentally distinct tasks, such direct transfer is largely impractical. Our experimental findings further highlight that many methods effective in CC scenarios fail entirely in generation settings, sometimes even yielding adverse effects. This is primarily due to the substantial knowledge disparity across tasks, which causes the diffusion generator to collapse, a phenomenon we refer to as generative collapse. In this paper, we address this gap by developing a theoretical framework to model the task transition process of diffusion models.

\section{Continual Consistency Diffusion}
In our standardized CDG pipeline, we consider a sequence of non-stationary tasks \(\{\mathcal{T}_k\}_{k=1}^K\), where each task \(\mathcal{T}_k\) has a distinct data distribution  \(p^k(x_0)\) and corresponding label distribution \(p^k(y | x_0)\). The forward diffusion process for each task is governed by an SDE: \(dx_t^{k} = f_k(x_t^{k},t)dt + g_k(t)dw_t\), where \(x_t^{k} \in \mathbb{R}^d\) represents the diffused samples at time \(t\) under task \(\mathcal{T}_k\), and $g_k(t)$ controls time-dependent noise for each task. To build on the derivations in Appendix and enhance the preservation of shared knowledge across tasks, we employ a direct rehearsal buffer \(\mathcal{B}^{real}_k = \{ \hat{x}_0, \hat{y} \sim p^j(x_0, y) \}_{j=1}^{k-1}\), \footnote{We avoid generative replay as in DDGR \cite{gao2023ddgr}, since it requires synthesizing past samples for every training batch, leading to a 2–3× increase in optimization time and rendering it impractical despite its performance gains.} which stores a limited set of real samples within a fixed storage budget \(C\), following common practice in CL research such as \cite{litensor,wanunderstanding,wang2025cut}. This memory mechanism, both in its theoretical formulation and practical implementation, forms the core foundation for the retention of shared knowledge across tasks.

Building on these foundations, we now present a detailed exposition of our CCD optimization framework’s theory and loss formulation within the standardized CDG pipeline.

\subsection{Theoretical Foundation}
A central challenge in CDG is formalizing the interaction of task-specific generative processes through shared SDE dynamics \cite{song2021scorebased}, as in Equation~\ref{cross_sde} in the Appendix. Existing empirical approaches \cite{smith2024continual, zhao2024invertavatar, zhang2024clog} mainly adapt solutions from continual classification methods, but lack rigorous theoretical guarantees, often leading to GCF or rigid fixation. Our analysis begins by establishing fundamental bounds on cross-task knowledge retention, which are essential for systematic CDG. Through rigorous derivation, we establish Theorem~\ref{thm:cross-task-diffusion}, which lays the theoretical foundation for subsequent innovations.

\begin{theorem}[Cross-Task Diffusion Evolution Bound] \label{thm:cross-task-diffusion} Let \(\mathcal{T}_i\) and \(\mathcal{T}_j\) represent two tasks in CDG, each characterized by distinct data distributions \(p(x_0)\) and \(q(x_0)\), along with their respective conditional prior distributions \(p(y)\) and \(q(y)\). The diffused processes for these tasks evolve over time as \(\{p(x_t)\}_{t=0}^T\) and \(\{q(x_t)\}_{t=0}^T\). Assume that for all \(x_0, y, t\), the conditional probability distributions of the two tasks satisfy \(p_t(x_t | x_0, y) = q_t(x_t | x_0, y)\). Here, \(p_t(x_t | x_0, y)\) refers to the distribution \(p(x_t | x_0, y, t)\), and similarly, \(q_t(x_t | x_0, y)\) denotes \(q(x_t | x_0, y, t)\). Furthermore, let \(\epsilon_\theta^p\) and \(\epsilon_\theta^q\) represent the time-dependent score approximators, or noise estimators, for tasks \(\mathcal{T}_i\) and \(\mathcal{T}_j\). Under mild assumptions, we expect that the gradients of the mean functions \(\mu(x_t, t)\) and \(\nu(x_t, t)\) align, such that \(\nabla_{x_t}\mu(x_t, t) \approx \nabla_{x_t}\nu(x_t, t)\), as noted in \cite{yu2020gradient}. Additionally, it is assumed that the variance \(\sigma^2_t\) at any given time \(t\) remains consistent across both tasks. Lastly, we assume that the evolving state \(x_t\) does not influence the label \(y\), meaning the label is independent of the diffusion process at any given time step. These conditions enable the potential retention and transfer of knowledge between tasks, leading to the derivation of an optimization upper bound for their interaction. \footnote{For the detailed proof, please see Appendix.}

There exist constants \(\{\kappa, \lambda, \eta\} \subset \mathbb{R}_{>0}\) such that the inter-task discrepancy is uniformly bounded:
\begin{equation} \label{diffusion_transfer_loss}
    \mathcal{L}_{UB}
    \;=\;
    \kappa \mathcal{L}_{IKC}
    \;+\; \lambda\,\mathcal{L}_{UKC}
    \;+\; \eta\,\mathcal{L}_{PKC},
\end{equation}
where
\begin{equation} \label{lsde}
    \mathcal{L}_{IKC}
    \;=\;
    \epsilon_\theta^{q}(x_t, y, t) \;-\; \epsilon_\theta^{p}(x_t, y, t),
\end{equation}
\begin{equation}
\mathcal{L}_{UKC}
    \;=\;
    \frac{\bar{\alpha}_t^2}{\bar{\beta}_t^2}\,\bigl[\mu_\theta(x_t, t) - \nu_\theta(x_t, t)\bigr], 
\end{equation}
\begin{equation}
\mathcal{L}_{PKC} \label{pkc}
    \;=\;
    \frac{\bar{\alpha}_t}{\bar{\beta}_t}
    \mathbb{E}_{p_t(x_0 | x_t)}
    \Bigl[D_{\mathrm{KL}}\bigl(p_t(y | x_0) \,\big\|\, q_t(y | x_0)\bigr)\Bigr].
\end{equation}

In particular, \(\mathcal{L}_{UB}\) encapsulates three components: the inter-task knowledge consistency (\(\mathcal{L}_{IKC}\)), the unconditional knowledge consistency (\(\mathcal{L}_{UKC}\)), and the prior knowledge consistency (\(\mathcal{L}_{PKC}\)). Minimizing \(\mathcal{L}_{UB}\) aligns the reverse-time diffusion gradients between tasks \(\mathcal{T}_i\) and \(\mathcal{T}_j\), thereby allowing the two tasks to retain as much shared knowledge as possible during the SDE optimization process.
\end{theorem}


\subsection{Basic Diffusion Model Training} 
To ensure the optimality of the score estimators \(\epsilon_\theta^p\) and \(\epsilon_\theta^q\) within \(\mathcal{L}_{IKC}\) for effective subsequent retention and transfer, we define the fundamental training objective for each task \(\mathcal{T}_k\). Following the standard DDPM framework \cite{ho2020denoising,song2021scorebased}, the base objective for conditional generation minimizes the weighted \(L_2\) error between predicted and actual noise:
\begin{equation} \label{eq:combined_diff_loss}
    \mathcal{L}_{cond}^k = \mathbb{E}_{t \sim \mathcal{U}(0,T), \epsilon \sim \mathcal{N}(0,I)} \left[ \|\epsilon_\theta(\bar{\alpha}_t x_0 + \bar{\beta}_t \epsilon, t, y) - \epsilon\|_2^2 \right],
\end{equation}
where \(\bar{\alpha}_t\) and \(\bar{\beta}_t\) follow the DDPM variance schedule \cite{ho2020denoising}, and \(\epsilon\) is the standard Gaussian noise.

Building on this, to enhance the shared knowledge between different tasks, we incorporate the data from the rehearsal buffer into the training process. However, instead of random sampling, we concatenate the pairs \(<x_0, \hat{x}_0>\), which also facilitates the implementation of Equations \ref{lsde} to \ref{pkc}. Therefore, in CDG pipeline, the fundamental composite training objective for diffusion models can be expressed as:
\begin{equation} \label{eq:cont_diff_loss}
\mathcal{L}_{base}^k = \mathcal{L}_{cond}^k + \mathbb{E}_{(\hat{x}_0, \hat{y}) \sim \mathcal{B}_{k}, t, \epsilon} \left[ \|\epsilon_\theta(\bar{\alpha}_t \hat{x}_0 + \bar{\beta}_t \epsilon, t, \hat{y}) - \epsilon\|_2^2 \right].
\end{equation}

This formulation theoretically ensures that the diffusion model maintains effective performance across evolving task distributions, thereby providing a stable optimization trajectory and establishing a reliable lower bound for \(\mathcal{L}_{IKC}\).

\subsection{Inter-task Knowledge Consistency} 
\label{sec:ised}
The inter-task knowledge consistency loss \(\mathcal{L}_{IKC} = \epsilon_\theta^{q} - \epsilon_\theta^{p}\) measures, for each sample $x$ and diffusion step $t$, the output and parameter difference between the estimator trained on task $q$ and the estimator retained from task $p$, thereby directly capturing their divergence across tasks. 
In CC tasks, L2 regularization is commonly employed to prevent excessive variations in parameters and outputs. However, in CDG, this approach may lead to catastrophic degradation, severely impairing the model’s generative capability, as demonstrated in Table~\ref{tab:results}. To circumvent this limitation, we introduce a new knowledge retention strategy, drawing inspiration from \cite{hinton2015distilling}, where the previously learned score estimator \(\epsilon_\theta^p\) acts as a teacher to guide the adaptation of \(\epsilon_\theta^q\). In contrast to conventional knowledge distillation techniques \cite{moslemi2024survey}, which primarily manipulate class probability distributions, our approach capitalizes on the stochastic gradients governed by the reverse-time SDE. This sophisticated formulation not only facilitates an exceptionally seamless and cohesive knowledge retention but also profoundly mitigates the propensity for GCF, thereby preserving generative fidelity across successive tasks.

Let \(\mathcal{M}_{k-1}\) denote the frozen diffusion model for task \(\mathcal{T}_{k-1}\), parameterized by \(\theta_{k-1}\). For a new task \(\mathcal{T}_{k}\), we seek to adapt \(\theta_{k}\) while preserving the score-matching capability on prior tasks. To achieve this, we minimize the \emph{\textbf{Bregman divergence}} \cite{siahkamari2020learning} between the score distributions of \(\mathcal{M}_{k-1}\) and \(\mathcal{M}_{k}\) over a shared noise manifold. Specifically, given the current samples \( (x_t^{k}, y^k) \sim p^{k} \) from task \(\mathcal{T}_{k}\) and the replayed samples \( (\hat{x}_t^{k}, \hat{y}^k) \sim \mathcal{B}_{k}\), the $\mathcal{L}_{IKC}$ is defined as:
\begin{equation} \label{eq:ised_loss}
\footnotesize
\begin{aligned}
    \mathcal{L}_{\mathrm{IKC}} = \mathbb{E}_{\hat{x}_t^{k},\hat{y}^{k},x_t^{k},y^{k}, t} 
    \Big[D_{\varphi} \big( \epsilon_{\theta_{k-1}}(\hat{x}_t^{k}, \hat{y}^{k}, t) \parallel \epsilon_{\theta_{k}}(x_t^{k}, y^{k}, t) \big) \Big],
\end{aligned}
\end{equation}
where \(D_{\varphi}\) is an adaptation via local Bregman divergence minimization with curvature matrix \(\varphi\).

Crucially, we generalize the conventional squared \(\ell_2\) distance to a curvature-aware Bregman divergence, defined via a locally-varying positive definite matrix \(\varphi\) that reflects the geometry of the score function landscape.
\begin{equation} \label{eq:mahalanobis_div}  
D_{\varphi}(u \parallel v) = \frac{1}{2}(u - v)^\top \varphi(\hat{x}_t^{k}, \hat{y}^{k}, t)(u - v).
\end{equation}    

The preconditioner \(\varphi(\hat{x}_t^{k}, \hat{y}^{k}, t)\) is derived from the data space metric of \(\mathcal{M}_{k-1}\):  
\begin{equation}\label{eq:fisher_preconditioner}
\begin{split}
\varphi(\hat{x}_t^{k}, \hat{y}^{k}, t) = \mathbb{E} \Bigl[
\nabla_{\hat{x}_t^{k}}\log \epsilon_{\theta_{k-1}}(\hat{x}_t^{k}|\hat{y}^k, t)\, \times \\
\nabla_{\hat{x}_t^{k}}\log \epsilon_{\theta_{k-1}}(\hat{x}_t^{k}|\hat{y}^k, t)^\top
\Bigr].
\end{split}
\end{equation}

By aligning the divergence metric with the gradient information, which captures the curvature of the teacher model's parameters \(\mathcal{M}_{k-1}\), the student model \(\mathcal{M}_{k}\) maintains high consistency with its built-in knowledge, effectively reducing the knowledge gap between the two models in Eq.~\ref{lsde}.

\subsection{Unconditional Knowledge Consistency} \label{sec:ugd}  
Building on the inter-task model alignment, we now instantiate the $\mathcal{L}_{UKC}$ term from Theorem~\ref{thm:cross-task-diffusion}, which enforces consistency in the mean of unconditional sample embeddings and reverse-time denoising trajectories. This component serves as a bridge between theoretical guarantees and practical implementation by explicitly aligning the mean functions of reverse processes across tasks.

Deriving task-specific reverse mean functions presents a fundamental challenge. Given current instances \((x_0^{k}, y^k) \sim (p^k \cup \mathcal{B}_{k})\) \footnote{In $\mathcal{L}_{UKC}$, the primary objective is to preserve geometric constraints and retain knowledge across both historical and current data. In the ensuing subsections, we reinforce this continuity by integrating historical data, thereby treating the current samples as a comprehensive amalgamation of both past and present information.} and buffered historical samples \((\hat{x}_0^{k}, \hat{y}^{k}) \sim \mathcal{B}_{k}\), direct computation of \(\mathbb{E}[x_t]\) remains ill-posed due to the artificial noise-driven construction of \(x_t\) in diffusion frameworks. This arises from the intrinsic semantic mismatch in perturbed diffused states \(x_t\) across training phases, rendering naive trajectory averaging and constraint imposition ineffective for gradient-based optimization. To address this, we devise an indirect mean constraint through symbiotic knowledge transfer between the frozen teacher \(\mathcal{M}_{k-1}\) and adaptive student \(\mathcal{M}_k\). By reparameterizing \(\mathbb{E}[x_{t-1}]\), we enable full gradient backpropagation while enforcing coherent diffused space constraints. Crucially, label-marginalized computation ensures these constraints govern unconditional generation fidelity. The reverse process mean functions for both historical and current models are derived analytically within the shared data using their respective noise prediction networks:  
\begin{equation} \label{eq:multi_task_means}
    \begin{aligned}    
    \mu_{\theta_k}(x_{t - 1}, t - 1) &= \frac{1}{\sqrt{\alpha_t}}x_t^k - \frac{1-\alpha_t}{\sqrt{\alpha_t(1-\bar{\alpha}_t)}}\epsilon_{\theta_k}(x_t^k, t), \\
    \mu_{\theta_{k-1}}(\hat{x}_{t - 1}, t - 1) &= \frac{1}{\sqrt{\alpha_t}}\hat{x}_t^k - \frac{1-\alpha_t}{\sqrt{\alpha_t(1-\bar{\alpha}_t)}}\epsilon_{\theta_{k-1}}(\hat{x}_t^k, t).
    \end{aligned}  
\end{equation}
Here, $\mu_{\theta_{k-1}}(\cdot)$ and $\mu_{\theta_k}(\cdot)$ are the posterior means predicted by the previous and current models, respectively. $\alpha_t = 1 - \beta_t$ and $\bar{\alpha}_t = \prod_{s=1}^t \alpha_s$ are standard DDPM forward coefficients.

To enforce unconditional mean consistency across incremental adaptations, we formulate $\mathcal{L}_{UKC}$ as a time-weighted divergence between these mean estimates:  
\begin{equation} \label{eq:ugd_final}
    \mathcal{L}_{UKC} = \frac{\bar{\alpha}_t^2}{1-\bar{\alpha}_t^2} \left\| \mu_{\theta_k}(x_{t - 1}, t - 1) - \mu_{\theta_{k-1}}(\hat{x}_{t - 1}, t - 1) \right\|_2^2.  
\end{equation}  

The weighting term \(\frac{\bar{\alpha}_t^2}{1-\bar{\alpha}_t^2}\) emphasizes alignment during semantically critical mid-diffusion phases, where latent structures transition between noise and meaningful representations. By penalizing deviations in denoising trajectories and the mean of unconditional sample embeddings, this loss enforces constraints that preserve the manifold topology of historical data within the evolving student model \(\mathcal{M}_{k}\). This mechanism complements the instantaneous model matching strategy outlined in $\mathcal{L}_{\mathrm{IKC}}$, ensuring both local knowledge coherence and global structural fidelity.

\subsection{Prior Knowledge Consistency} \label{sec:gscd}
The prior knowledge consistency loss \(\mathcal{L}_{PKC}\) preserves the shared semantic prior, instantiated as label information in this work, across the original samples. We realize it with a label regressor that gauges the semantic proximity of \(\hat{x}_0\) and \(x_0\) in label space, following multi‑domain alignment strategies in zero‑shot learning \cite{hwang2014unified,ni2019dual,li2023vs,duan2024visual,zhang2024s3a}. Because the regressor is trained only for similarity, not classification accuracy, it avoids label‑collapse and readily generalizes to textual or other modality priors in diffusion models.
Let $\mathcal{M}_{k}$ denote the current task model with its task‑adaptive regressor $h_\phi^{k}$, and let the frozen regressor $h_\phi^{<k}$ from $\mathcal{M}_{k-1}$ preserve earlier semantics. Although one could generate $\hat{x}'_0 = \mathcal{M}_{k-1}(\epsilon, \hat{y}, T)$ via $p_t(x_0|x_t)$, this generative replay is costly and produces images almost identical to stored ones. Hence, we simply draw $\hat{x}_0$ from the buffer $\mathcal{B}_{k}$ and pair it with the current sample $x_0$ to compute $\mathcal{L}_{PKC}$.
\begin{equation}
\mathcal{L}_{PKC} = \frac{\bar{\alpha}_t}{\bar{\beta}_t} \mathbb{E}_{\hat{x}_0 \sim \mathcal{B}_{k}} \left[ D_{\mathrm{KL}} \left( h_\phi^{< k}(y|\hat{x}_0) \parallel h_\phi^{k}(y|x_0) \right) \right].
\end{equation}

This objective enforces semantic consistency between past and current samples by extracting shared prior knowledge, thereby curbing label‑space drift. Ultimately, together with the other two optimizations, diffusion models achieve continuous alignment in model parameters, unconditional mean distributions, and the prior space, effectively preserving shared knowledge across continual generation tasks, thereby achieving improved long-term generative performance.

\section{Experiments}
\subsection{Datasets and Benchmark Characteristics} \label{datasets}
We conduct comprehensive experiments on five representative vision benchmarks spanning diverse domains.  
\textbf{MNIST} \cite{lecun1998gradient} consists of 60,000 training and 10,000 test images spanning 10 handwritten digit classes. In our setting, the dataset is partitioned into 5 disjoint tasks, each comprising a subset of the digit classes. 
\textbf{OxfordPets} \cite{parkhi2012cats} comprises 7,349 RGB images across 37 classes. To accommodate the CDG pipeline, we split it into five tasks, excluding two classes to maintain an equal number of categories per task.
\textbf{CIFAR-100} \cite{krizhevsky2009learning} comprises 100 object categories, each containing 600 images. The dataset poses significant challenges for our CCD optimization framework due to its low image resolution and minimal knowledge overlap across tasks. It is partitioned into 10 tasks, each with 10 classes.  
\textbf{Flowers102} \cite{nilsback2008automated} consists of 8,189 images from 102 flower species with significant intra-class variance. It is divided into 10 tasks for fine-grained generation evaluation.  
\textbf{CUB-200-2011} \cite{wah2011caltech} offers 11,788 bird images from 200 species with subtle morphological variations, split into 10 tasks.  
To ensure fairness and consistent GPU memory consumption, all images are uniformly resized to \( 32 \times 32 \).

\subsection{CDG Pipeline}
We present an end-to-end pipeline as a rigorous and standardized framework for diffusion-based continual generation methods. Centered on a unified UNet2D diffusion backbone (see Appendix Figure.~\ref{frameworks}), the model is trained using the standard DDPM \cite{ho2020denoising} formulation with 1000 denoising steps and an MSE loss derived from Gaussian noise prediction. Optimization employs the Adam optimizer with a batch size of 200 and a fixed learning rate of $1 \times 10^{-3}$. During inference, a configurable DDIM \cite{song2020denoising} scheduler with 50 sampling steps is used. For evaluation, 2048 samples are generated and uniformly distributed across classes to ensure metric fairness. All experiments are conducted on NVIDIA A800 GPUs under consistent computational settings. To rigorously assess GCF, we introduce two complementary metrics: Mean Fidelity (MF), capturing terminal-state generative quality, and Incremental Mean Fidelity (IMF), quantifying temporal stability. Given a temporally ordered task manifold $\{\mathcal{T}_k\}_{k=1}^K$, MF is defined as $\mathrm{MF} = \mathbb{E}_{k \sim [1, K]}[d_\mathcal{M}(p^k_{\text{real}} \parallel p^k_{\text{gen}})]$, where $d_\mathcal{M}$ denotes a generative quality metric (e.g., FID \cite{heusel2017gans}). IMF extends this by aggregating performance over time: $\mathrm{IMF} = \mathbb{E}_{k \sim [1, K]}[\mathbb{E}_{i \leq k}[d_\mathcal{M}(p^i_{\text{real}} \parallel p^i_{\text{gen}})]]$, reflecting expected stability under continual updates. This dual-metric formulation enables precise characterization of both endpoint fidelity and generative consistency across task sequences.

\begin{table*}[t]
    \centering
    \small
    \renewcommand{\arraystretch}{1.0}
    \setlength{\tabcolsep}{1.1mm}
    \begin{tabular}{@{}c c cc cc cc cc cc|>{\centering\arraybackslash}p{8mm} >{\centering\arraybackslash}p{8mm} >{\centering\arraybackslash}p{8mm} >{\centering\arraybackslash}p{8mm}@{}}
        \toprule
        \rowcolor{gray!15}
        \textbf{Method} & \textbf{Venue}
        & \multicolumn{2}{c}{\textbf{MNIST-5T}} 
        & \multicolumn{2}{c}{\textbf{OxfordPets-5T}}
        & \multicolumn{2}{c}{\textbf{CIFAR100-10T}} 
        & \multicolumn{2}{c}{\textbf{Flowers102-10T}} 
        & \multicolumn{2}{c}{\textbf{CUB200-10T}}
        & \multicolumn{2}{c}{\textbf{Weighted Avg}} \\
        \cmidrule(lr){3-4} \cmidrule(lr){5-6} 
        \cmidrule(lr){7-8} \cmidrule(lr){9-10} \cmidrule(lr){11-12} \cmidrule(lr){13-14}
        & & MF↓ & IMF↓ & MF↓ & IMF↓ & MF↓ & IMF↓ & MF↓ & IMF↓ & MF↓ & IMF↓ & MF↓ & IMF↓ \\
        \midrule
        Non-CL & -- 
        & 4.99 & 4.99   & 224.26 & 224.26 & 65.97 & 65.97   & 21.16 & 21.16   & 35.80 & 35.80
        & 1.00 & 1.00 \\
        \midrule
        \rowcolor{gray!10} \multicolumn{14}{c}{\textbf{Rehearsal-free Methods}} \\
        LwF & \textit{TPAMI} (2017) 
        & 62.83 & 58.98   & 288.35 & 283.91 & 114.22 & 103.40   & 102.49 & 117.79   & 157.70 & 111.07
        & 4.97 & 4.66 \\
        EWC & \textit{ICLR} (2017) 
        & 119.29 & 176.59 & 307.85 & 333.72 & 110.07 & 141.78   & 104.91 & 114.35   & 165.19 & 163.08
        & 7.30 & 9.80 \\
        L2 & -- 
        & 105.34 & 67.66 & 272.00 & 267.30 & 134.25 & 129.48   & 153.53 & 132.70   & 125.47 & 108.61
        & 7.02 & 5.20 \\
        SI & \textit{PMLR} (2017) 
        & 85.39 & 66.95 & 269.22 & 267.75 & 145.06 & 132.67   & 148.00 & 136.80   & 142.90 & 132.01
        & 6.30 & 5.35 \\
        MAS & \textit{ECCV} (2018) 
        & 291.70 & 304.40 & 201.77 & 222.17 & 146.77 & 169.19   & 104.99 & 317.51   & 128.66 & 154.84
        & 14.03 & 16.78 \\
        C-LoRA & \textit{TMLR} (2024) 
        & 106.71 & 80.63 & 457.49 & 442.48 & 134.58 & 131.47   & 369.62 & 363.88   & 142.85 & 129.43
        & 9.38 & 8.19 \\
        \midrule
        \rowcolor{gray!10} \multicolumn{14}{c}{\textbf{Storage Rehearsal Methods (512 buffer)}} \\
        ER & \textit{NeurIPS} (2017) 
        & 97.57 & 69.38 & \underline{287.12} & \textbf{276.05} & \textbf{94.19} & \textbf{89.30}     & \underline{88.36} & 94.06     & 130.57 & \underline{135.07}
        & 6.02 & 4.94 \\
        A-GEM & \textit{ICLR} (2019) 
        & 94.67 & 66.00 & 289.42 & \underline{279.36} & 96.85 & 90.80     & 89.41 & \underline{92.59}     & 121.73 & 139.66
        & 5.87 & \underline{4.83} \\
        LwF + ER & - 
        & 89.46 & 105.41 & 306.58 & 290.70 & 112.80 & 113.97   & 245.86 & 319.55   & 119.10 & 152.15
        & 7.19 & 8.70 \\
        DCM & \textit{CVPR} (2024) 
        & \textbf{63.06} & \textbf{43.74} & 290.77 & 282.36 & 100.06 & 96.91   & 187.25 & 259.39   & \textbf{112.62} & 143.58
        & \underline{5.49} & 5.55 \\
        \rowcolor{cyan!15}
        CCD + ER & \textit{Ours} 
        & \underline{79.76} & \underline{57.63} & \textbf{285.81} & 279.87 & \underline{96.34} & \underline{90.14}     & \textbf{86.11} & \textbf{92.18}     & \underline{121.15} & \textbf{101.42}
        & \textbf{5.23} & \textbf{4.27} \\
        \midrule
        \rowcolor{gray!10} \multicolumn{14}{c}{\textbf{Storage Rehearsal Methods (2560 buffer)}} \\
        ER & \textit{NeurIPS} (2017) 
        & 73.60 & 56.95 & 288.19 & \underline{280.73} & 103.38 & \underline{98.55}    & 67.89 & 80.33     & 373.50 & 353.32
        & 6.25 & 5.56 \\
        A-GEM & \textit{ICLR} (2019) 
        & 69.77 & \underline{54.32} & 287.59 & 280.86 & 104.58 & 99.43    & \underline{66.36} & \underline{79.79}     & \underline{103.78} & \underline{114.93}
        & \underline{4.58} & \underline{4.13} \\
        LwF + ER & - 
        & 75.42 & 115.63 & 296.24 & 286.72 & 106.63 & 107.52   & 212.12 & 316.77   & \textbf{95.64} & 144.92
        & 6.15 & 9.02 \\
        DCM & \textit{CVPR} (2024) 
        & \textbf{61.11} & 108.93 & \underline{285.79} & 281.81 & \textbf{96.24} & \textbf{89.15}     & 199.57 & 264.15   & 107.00 & 138.27
        & 5.48 & 8.16 \\
        \rowcolor{cyan!15}
        CCD + ER & \textit{Ours} 
        & \underline{67.48} & \textbf{50.94} & \textbf{237.27} & \textbf{254.41} & \underline{101.51} & 98.68    & \textbf{65.31} & \textbf{79.70}     & 106.93 & \textbf{84.11}
        & \textbf{4.44} & \textbf{3.79} \\
        \midrule
        \rowcolor{gray!10} \multicolumn{14}{c}{\textbf{Storage Rehearsal Methods (5120 buffer)}} \\
        ER & \textit{NeurIPS} (2017) 
        & 60.40 & 135.19 & 276.00 & 274.99 & 103.34 & 99.39    & 67.19 & 80.23     & 135.98 & 164.87
        & 4.38 & 7.64 \\
        A-GEM & \textit{ICLR} (2019) 
        & 68.59 & 107.24 & \underline{272.63} & \underline{273.84} & 103.87 & 100.49   & \underline{65.39} & \underline{79.71}     & \underline{70.33}  & \underline{99.21}
        & \underline{4.32} & 6.04 \\
        LwF + ER & - 
        & 57.25 & 72.87 & 277.31 & 278.84 & 106.13 & \textbf{97.61}   & 212.12 & 316.77   & 88.88 & 142.35
        & 5.37 & 7.25 \\
        DCM & \textit{CVPR} (2024) 
        & \underline{50.45} & \textbf{25.48} & 285.79 & 279.52 & \textbf{101.62} & 100.79   & 199.57 & 264.15   & 110.10 & 119.37
        & 5.09 & \underline{4.74} \\
        \rowcolor{cyan!15}
        CCD + ER & \textit{Ours} 
        & \textbf{48.97} & \underline{40.08} & \textbf{260.86} & \textbf{244.48} & \underline{102.09} & \underline{98.33}    & \textbf{64.97} & \textbf{79.62}     & \textbf{67.00}  & \textbf{61.37}
        & \textbf{3.49} & \textbf{3.22} \\
        \bottomrule
    \end{tabular}
    \vspace{-1ex}
    \caption{Performance comparison across datasets and buffer sizes. Top two per buffer size are marked in bold and underline. Weighted Averages are computed as the normalized mean of MF↓ and IMF↓ using the Non-CL scores as the baseline.}
    \vspace{-4ex}
    \label{tab:results}
\end{table*}

\subsection{Performance Comparison}
In this subsection, we benchmark representative baselines under standardized hardware and software settings. These include CC methods, LwF \cite{li2017learning}, EWC \cite{kirkpatrick2017overcoming}, SI \cite{zenke2017continual}, MAS \cite{aljundi2018memory}, ER \cite{lopez2017gradient}, and A-GEM \cite{chaudhry2018efficient}, as well as generative CL approaches such as C-LoRA \cite{smith2024continual} and DCM \cite{ye2024online}. The implementation details are provided in the Appendix.
Although buffer-based CL methods like DDGR \cite{gao2023ddgr}, TD \cite{litensor}, AM \cite{wanunderstanding}, and CUTER \cite{wang2025cut} perform well in CC tasks, they rely on sample embeddings or prototypes, an assumption incompatible with diffusion models, whose latent space consists of isotropic Gaussian noise. Therefore, we do not consider them in this work. We also exclude approaches such as \cite{zajkac2023exploring, masip2023continual, cheng2024semi} that depend on pretrained diffusion backbones and generative replay buffers, as they violate the controlled comparison protocol. These exclusions highlight the methodological gap between continual classification and diffusion-based generation, motivating the treatment of our CDG pipeline as a standalone research subject.

To ensure fair comparison, we apply our CCD optimization framework atop the ER buffer. As shown in Table \ref{tab:results}, CCD consistently outperforms most baselines across varying buffer sizes and evaluation metrics, with its advantage becoming more pronounced as memory increases. This trend is especially evident on MNIST-5T, Flowers102-10T, and CUB200-10T, where CCD reduces MF and IMF scores by over 50\% relative to ER in some cases, indicating substantial mitigation of GCF. Compared to the state-of-the-art DCM, CCD demonstrates superior resilience to catastrophic collapse, particularly on fine-grained datasets such as Flowers102-10T and CUB200-10T, where it achieves lower IMF scores. Moreover, on OxfordPets-5T, where most buffer-based methods fail to yield any meaningful generative improvement, our method, at a buffer size of 2560, nearly matches the performance of a Non–CL upper bound. Finally, while DCM performance saturates with larger buffers, CCD continues to drive metrics lower, highlighting its scalability.

Nevertheless, we identify a key limitation: since CCD is designed to preserve shared knowledge across tasks, its effectiveness diminishes on coarse-grained datasets like CIFAR100-10T, where semantic overlap is minimal. This finding highlights a key open challenge for future research: \textit{developing effective strategies for knowledge retention and propagation under conditions of minimal cross-task overlap.}

\begin{figure}[h]
    \centering
    \includegraphics[scale=0.16]{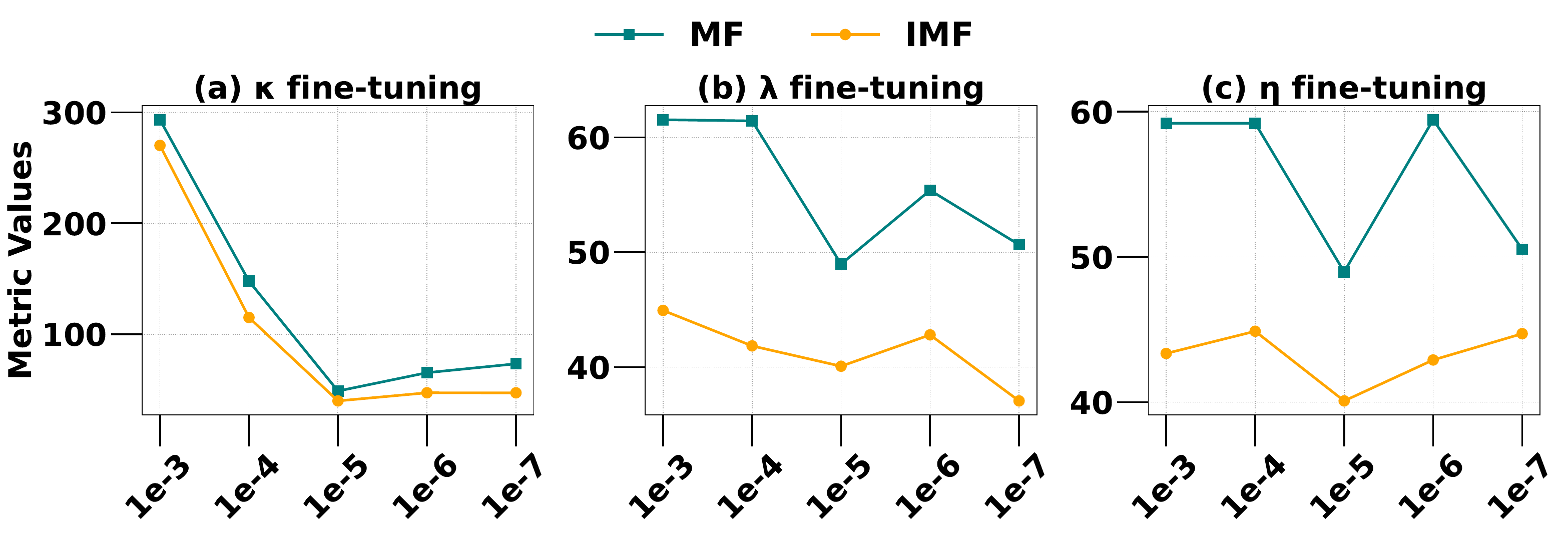}
    \vspace{-4ex}
    \caption{Hyperparameter sensitivity analysis on MNIST, illustrating the effects of fine-tuning \(\kappa\), \(\lambda\), and \(\eta\).}
    \label{Fig2}
    \vspace{-4ex}
\end{figure}

Furthermore, to provide a more comprehensive evaluation, we include the various perceptual metrics in the Appendix.


\subsection{Ablation Studies}
In this subsection, we perform a sensitivity analysis on Eq.~\ref{diffusion_transfer_loss} using MNIST and an ER-5120 buffer strategy. As shown in Figure.~\ref{Fig2}, CDG performance is strongly influenced by three key hyperparameters: the model consistency coefficient $\kappa$, the unconditional generation consistency weight $\lambda$, and the prior knowledge consistency coefficient $\eta$. The optimal MF and IMF scores are observed when $\kappa = 1 \times 10^{-5}$. Increasing $\kappa$ leads to overreliance on prior models, while decreasing it weakens consistency and retention. Similarly, $\lambda = 1 \times 10^{-5}$ yields the best results, whereas higher values overly constrain the unconditional mean, limiting denoising flexibility. The prior knowledge consistency coefficient $\eta$ also peaks at $1 \times 10^{-5}$, effectively preserving semantic structure across tasks. These results highlight the importance of carefully balancing temporal, generative, and semantic consistency. Each component plays a distinct role in mitigating GCF and their joint calibration is essential for the CCD performance.

\begin{figure}[!t]
    \centering
    \includegraphics[scale=0.14]{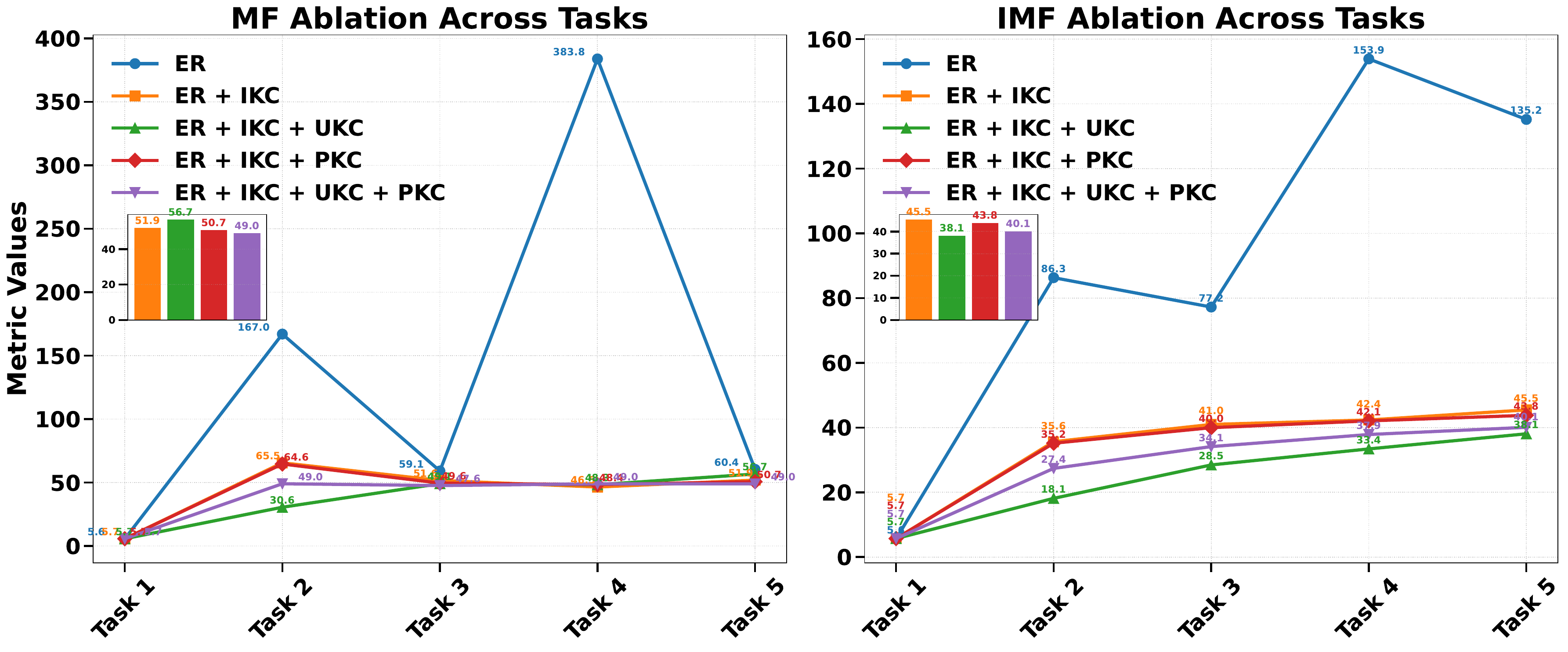}
    \vspace{-4ex}
    \caption{Ablation studies on MNIST-5T (ER buffer 5120).}
    \label{Fig3}
    \vspace{-1ex}
\end{figure}

In addition, to further disentangle the individual contributions of each loss term, we also conduct controlled ablation studies on MNIST and an ER-5120 buffer strategy, summarized in Figure.~\ref{Fig3}. The baseline model (MF: 60.40, IMF: 135.19) suffers from severe forgetting and collapse, highlighting the instability of unconstrained diffusion across tasks. Adding $\mathcal{L}_{IKC}$ declines both MF and IMF, confirming the role of inter-task model consistency in stabilizing forward dynamics. Incorporating $\mathcal{L}_{UKC}$ further enhances performance by preserving reverse-time denoising through unconditional mean alignment. Replacing $\mathcal{L}_{UKC}$ with $\mathcal{L}_{PKC}$ yields lower MF but slightly higher IMF, indicating that label-space alignment is more effective for semantic retention. Combining all three losses achieves the best results (MF: 48.97, IMF: 40.08), validating that multi-level consistency is essential for mitigating forgetting in the CDG pipeline.


\subsection{Visualization}
To visualize how generative performance evolves in our CCD framework, we present a qualitative case study on CUB200-10T with an ER-5120 buffer strategy. Using the model trained through the final task, we generate $32\times32$ images via a DDIM scheduler, sampling equally from Tasks 0, 4, and 8 (Figure.~\ref{Fig3}). Under standard ER replay, outputs collapse into near-pure noise, evidencing severe GCF. Introducing IKC and PKC losses progressively restores coarse contours, and the full CCD framework produces fully resolved images that permit unambiguous species identification. This stepwise reconstruction vividly illustrates CCD’s capacity to retain and consolidate knowledge across sequential tasks. For comprehensive visual examples, please see the Appendix.

\subsection{Buffer Construction Strategies}
\begin{figure}[h]
    \centering
    \includegraphics[scale=0.95]{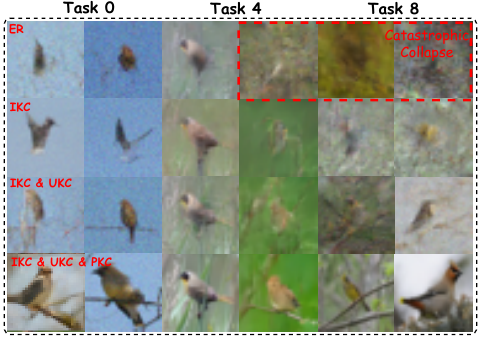}
    \vspace{-2ex}
    \caption{Ablation study showcasing the generated images for tasks 0, 4, and 8 on the CUB200-10T benchmark using the model trained up to the final task.}
    \label{Fig4}
    \vspace{-1ex}
\end{figure}

Moreover, we observe that the first-in-first-out (FIFO) strategy in ER \cite{lopez2017gradient} introduces substantial randomness, often prioritizing redundant samples at the cost of diversity. Although CCD partially alleviates the resulting performance variance, it compromises consistency across tasks. To address this limitation, we introduce a Hierarchical Diversity Buffer (HDB) architecture.

\noindent\textbf{HDB:}
To counteract the instability caused by FIFO-based sample replacement, HDB partitions memory into two components: a Candidate Pool for temporary intake and an Elite Repository for long-term storage of diverse exemplars. When the Candidate Pool reaches capacity, samples are assessed via similarity scoring using an exponential decay kernel, $S_{ij} = \exp(-\|\mathbf{h}_i - \mathbf{h}_j\|^2 / 2\alpha^2)$, where $\mathbf{h}_i$ and $\mathbf{h}_j$ denote prior regression embeddings from PKC. Samples with low average similarity (i.e., high diversity) are promoted to the Elite Repository. This progressive filtering maximizes coverage of the underlying data distribution while avoiding redundancy, leading to more stable and effective rehearsal-based training.

As shown in Table~\ref{tab:buffer_hdb}, HDB further improves CCD performance on coarse-grained datasets. However, we also observe certain adverse effects on fine-grained datasets. This limitation largely stems from the non-discriminative nature of intermediate outputs in diffusion models, making it challenging for HDB, and most CL methods, to effectively determine which samples should be retained. This insight points to a promising direction for future work: \textit{enhancing the discriminative quality of intermediate diffusion representations}.

\begin{table}
    \centering
    \fontsize{7pt}{7pt}\selectfont
    \renewcommand{\arraystretch}{1.2}
    \setlength{\tabcolsep}{4pt}
    \begin{tabular}{c c c c c c c}
        \toprule
        \rowcolor{gray!20}
        \multirow{2}{*}{\textbf{Method}} & \multicolumn{2}{c}{\textbf{HDB-512}} & \multicolumn{2}{c}{\textbf{HDB-2560}} & \multicolumn{2}{c}{\textbf{HDB-5120}} \\
        \cmidrule(lr){2-3} \cmidrule(lr){4-5} \cmidrule(lr){6-7}
        \rowcolor{gray!20}
        & MF ($\downarrow$) & IMF ($\downarrow$) & MF ($\downarrow$) & IMF ($\downarrow$) & MF ($\downarrow$) & IMF ($\downarrow$) \\
        \midrule
        MNIST-5T & 50.05 & 23.96 & 47.55 & 22.28 & 44.27 & 22.12 \\
        OxfordPets-5T & 287.21 & 271.76 & 288.65 & 279.51 & 286.64 & 275.56 \\
        CIFAR100-10T & 102.34 & 93.60 & 93.36 & 85.08 & 90.89 & 80.70 \\
        Flowers102-10T & 90.31 & 97.37 & 81.43 & 81.41 & 66.38 & 80.01 \\
        CUB200-10T & 140.60 & 100.52 & 142.66 & 99.20 & 140.77 & 99.77 \\
        \bottomrule
    \end{tabular}
    \vspace{-2ex}
    \caption{Performance of our CCD optimization framework across different datasets under varying HDB buffer sizes.}
    \label{tab:buffer_hdb}
    \vspace{-4ex}
\end{table}


\section{Conclusion}
\label{sec:conclusion}
This work presents a principled framework for our standardized CDG pipeline, grounded in stochastic calculus, to mitigate GCF in diffusion models. We conceptualize forgetting as a misalignment in cross-task SDE dynamics and introduce three key consistency constraints, inter-task, unconditional mean, and prior space, to promote stable knowledge retention. Building on these insights, we propose the CCD framework, which enforces multi-level consistency through hierarchical loss functions that preserve both the geometric and semantic integrity of generative trajectories. Experiments on diverse benchmarks demonstrate that CCD achieves state-of-the-art performance in both MF and IMF. By bridging static diffusion modeling with real-world streaming tasks, CCD lays a solid foundation for continual generation tasks.

\bibliography{aaai2026}

\begin{thebibliography}{58}
\providecommand{\natexlab}[1]{#1}

\bibitem[{Ali, Rossi, and Bertozzi(2025)}]{ali2025cfts}
Ali, M.; Rossi, L.; and Bertozzi, M. 2025.
\newblock CFTS-GAN: Continual Few-Shot Teacher Student for Generative Adversarial Networks.
\newblock In \emph{International Conference on Pattern Recognition}, 249--262. Springer.

\bibitem[{Aljundi et~al.(2018)Aljundi, Babiloni, Elhoseiny, Rohrbach, and Tuytelaars}]{aljundi2018memory}
Aljundi, R.; Babiloni, F.; Elhoseiny, M.; Rohrbach, M.; and Tuytelaars, T. 2018.
\newblock Memory aware synapses: Learning what (not) to forget.
\newblock In \emph{Proceedings of the European Conference on Computer Vision}, 139--154.

\bibitem[{Anderson(1982)}]{anderson1982reverse}
Anderson, B.~D. 1982.
\newblock Reverse-time diffusion equation models.
\newblock \emph{Stochastic Processes and their Applications}, 12(3): 313--326.

\bibitem[{Bruce et~al.(2024)Bruce, Dennis, Edwards, Parker-Holder, Shi, Hughes, Lai, Mavalankar, Steigerwald, Apps et~al.}]{bruce2024genie}
Bruce, J.; Dennis, M.~D.; Edwards, A.; Parker-Holder, J.; Shi, Y.; Hughes, E.; Lai, M.; Mavalankar, A.; Steigerwald, R.; Apps, C.; et~al. 2024.
\newblock Genie: Generative interactive environments.
\newblock In \emph{International Conference on Machine Learning}.

\bibitem[{Chaudhry et~al.(2019)Chaudhry, Ranzato, Rohrbach, and Elhoseiny}]{chaudhry2018efficient}
Chaudhry, A.; Ranzato, M.; Rohrbach, M.; and Elhoseiny, M. 2019.
\newblock Efficient Lifelong Learning with A-GEM.
\newblock In \emph{International Conference on Learning Representations}.

\bibitem[{Cheng et~al.(2024)Cheng, Liu, Long, Wu, He, and Wang}]{cheng2024semi}
Cheng, J.; Liu, Y.; Long, B.; Wu, Z.; He, L.; and Wang, T. 2024.
\newblock Semi-Process Noise Distillation for Continual Mixture-of-Experts Diffusion Models.
\newblock In \emph{2024 China Automation Congress (CAC)}, 2807--2812. IEEE.

\bibitem[{Duan et~al.(2024)Duan, Chen, Guo, Xie, Ding, and Wang}]{duan2024visual}
Duan, B.; Chen, S.; Guo, Y.; Xie, G.-S.; Ding, W.; and Wang, Y. 2024.
\newblock Visual--Semantic Graph Matching Net for Zero-Shot Learning.
\newblock \emph{IEEE Transactions on Neural Networks and Learning Systems}.

\bibitem[{Gao and Liu(2023)}]{gao2023ddgr}
Gao, R.; and Liu, W. 2023.
\newblock Ddgr: Continual learning with deep diffusion-based generative replay.
\newblock In \emph{International Conference on Machine Learning}, 10744--10763. PMLR.

\bibitem[{Gu et~al.(2020)Gu, Li, Gao, Chen, Wu, Cai, Wang, and Zhang}]{gu2020association}
Gu, Y.; Li, J.; Gao, Y.; Chen, R.; Wu, C.; Cai, F.; Wang, C.; and Zhang, Z. 2020.
\newblock Association: Remind Your GAN not to Forget.
\newblock \emph{arXiv preprint arXiv:2011.13553}.

\bibitem[{Heusel et~al.(2017)Heusel, Ramsauer, Unterthiner, Nessler, and Hochreiter}]{heusel2017gans}
Heusel, M.; Ramsauer, H.; Unterthiner, T.; Nessler, B.; and Hochreiter, S. 2017.
\newblock Gans trained by a two time-scale update rule converge to a local nash equilibrium.
\newblock \emph{Advances in neural information processing systems}, 30.

\bibitem[{Hinton(2015)}]{hinton2015distilling}
Hinton, G. 2015.
\newblock Distilling the Knowledge in a Neural Network.
\newblock \emph{arXiv preprint arXiv:1503.02531}.

\bibitem[{Ho, Jain, and Abbeel(2020)}]{ho2020denoising}
Ho, J.; Jain, A.; and Abbeel, P. 2020.
\newblock Denoising diffusion probabilistic models.
\newblock \emph{Advances in Neural Information Processing Systems}, 33: 6840--6851.

\bibitem[{Hwang and Sigal(2014)}]{hwang2014unified}
Hwang, S.~J.; and Sigal, L. 2014.
\newblock A unified semantic embedding: Relating taxonomies and attributes.
\newblock \emph{Advances in Neural Information Processing Systems}, 27.

\bibitem[{Jung et~al.(2023)Jung, Han, Bang, and Song}]{jung2023generating}
Jung, D.; Han, D.; Bang, J.; and Song, H. 2023.
\newblock Generating instance-level prompts for rehearsal-free continual learning.
\newblock In \emph{Proceedings of the IEEE/CVF International Conference on Computer Vision}, 11847--11857.

\bibitem[{Kirkpatrick et~al.(2017)Kirkpatrick, Pascanu, Rabinowitz, Veness, Desjardins, Rusu, Milan, Quan, Ramalho, Grabska-Barwinska et~al.}]{kirkpatrick2017overcoming}
Kirkpatrick, J.; Pascanu, R.; Rabinowitz, N.; Veness, J.; Desjardins, G.; Rusu, A.~A.; Milan, K.; Quan, J.; Ramalho, T.; Grabska-Barwinska, A.; et~al. 2017.
\newblock Overcoming catastrophic forgetting in neural networks.
\newblock \emph{Proceedings of the National Academy of Sciences}, 114(13): 3521--3526.

\bibitem[{Krizhevsky and Hinton(2009)}]{krizhevsky2009learning}
Krizhevsky, A.; and Hinton, G. 2009.
\newblock Learning multiple layers of features from tiny images.
\newblock Technical Report~0, University of Toronto, Toronto, Ontario.

\bibitem[{LeCun et~al.(1998)LeCun, Bottou, Bengio, and Haffner}]{lecun1998gradient}
LeCun, Y.; Bottou, L.; Bengio, Y.; and Haffner, P. 1998.
\newblock Gradient-based learning applied to document recognition.
\newblock \emph{Proceedings of the IEEE}, 86(11): 2278--2324.

\bibitem[{Li et~al.(2023)Li, Zhang, Bian, Qu, Xie, Shi, and Fan}]{li2023vs}
Li, X.; Zhang, Y.; Bian, S.; Qu, Y.; Xie, Y.; Shi, Z.; and Fan, J. 2023.
\newblock VS-Boost: Boosting Visual-Semantic Association for Generalized Zero-Shot Learning.
\newblock In \emph{IJCAI}, 1107--1115.

\bibitem[{Li et~al.(2025)Li, Zhou, Huang, Chen, Qiu, and Zhao}]{litensor}
Li, Y.; Zhou, G.; Huang, Z.; Chen, X.; Qiu, Y.; and Zhao, Q. 2025.
\newblock Tensor Decomposition Based Memory-Efficient Incremental Learning.
\newblock In \emph{International Conference on Machine Learning}.

\bibitem[{Li and Hoiem(2017)}]{li2017learning}
Li, Z.; and Hoiem, D. 2017.
\newblock Learning without forgetting.
\newblock \emph{IEEE Transactions on Pattern Analysis and Machine Intelligence}, 40(12): 2935--2947.

\bibitem[{Lopez-Paz and Ranzato(2017)}]{lopez2017gradient}
Lopez-Paz, D.; and Ranzato, M. 2017.
\newblock Gradient episodic memory for continual learning.
\newblock \emph{Advances in Neural Information Processing Systems}, 30.

\bibitem[{Lu et~al.(2022)Lu, Zhou, Bao, Chen, Li, and Zhu}]{lu2022dpm}
Lu, C.; Zhou, Y.; Bao, F.; Chen, J.; Li, C.; and Zhu, J. 2022.
\newblock Dpm-solver: A fast ode solver for diffusion probabilistic model sampling in around 10 steps.
\newblock \emph{Advances in Neural Information Processing Systems}, 35: 5775--5787.

\bibitem[{Masip et~al.(2023)Masip, Rodriguez, Tuytelaars, and van~de Ven}]{masip2023continual}
Masip, S.; Rodriguez, P.; Tuytelaars, T.; and van~de Ven, G.~M. 2023.
\newblock Continual learning of diffusion models with generative distillation.
\newblock \emph{arXiv preprint arXiv:2311.14028}.

\bibitem[{Moslemi et~al.(2024)Moslemi, Briskina, Dang, and Li}]{moslemi2024survey}
Moslemi, A.; Briskina, A.; Dang, Z.; and Li, J. 2024.
\newblock A survey on knowledge distillation: Recent advancements.
\newblock \emph{Machine Learning with Applications}, 18: 100605.

\bibitem[{Ni, Zhang, and Xie(2019)}]{ni2019dual}
Ni, J.; Zhang, S.; and Xie, H. 2019.
\newblock Dual adversarial semantics-consistent network for generalized zero-shot learning.
\newblock \emph{Advances in neural information processing systems}, 32.

\bibitem[{Nichol and Dhariwal(2021)}]{nichol2021improved}
Nichol, A.~Q.; and Dhariwal, P. 2021.
\newblock Improved denoising diffusion probabilistic models.
\newblock In \emph{International Conference on Machine Learning}, 8162--8171. PMLR.

\bibitem[{Nie et~al.(2025)Nie, Zhu, You, Zhang, Ou, Hu, Zhou, Lin, Wen, and Li}]{nie2025large}
Nie, S.; Zhu, F.; You, Z.; Zhang, X.; Ou, J.; Hu, J.; Zhou, J.; Lin, Y.; Wen, J.-R.; and Li, C. 2025.
\newblock Large language diffusion models.
\newblock \emph{arXiv preprint arXiv:2502.09992}.

\bibitem[{Nilsback and Zisserman(2008)}]{nilsback2008automated}
Nilsback, M.-E.; and Zisserman, A. 2008.
\newblock Automated flower classification over a large number of classes.
\newblock In \emph{2008 Sixth Indian Conference on Computer Vision, Graphics \& Image Processing}, 722--729. IEEE.

\bibitem[{Parisi et~al.(2019)Parisi, Kemker, Part, Kanan, and Wermter}]{parisi2019continual}
Parisi, G.~I.; Kemker, R.; Part, J.~L.; Kanan, C.; and Wermter, S. 2019.
\newblock Continual lifelong learning with neural networks: A review.
\newblock \emph{Neural Networks}, 113: 54--71.

\bibitem[{Parkhi et~al.(2012)Parkhi, Vedaldi, Zisserman, and Jawahar}]{parkhi2012cats}
Parkhi, O.~M.; Vedaldi, A.; Zisserman, A.; and Jawahar, C.~V. 2012.
\newblock Cats and dogs.
\newblock In \emph{Proceedings of the IEEE Conference on Computer Vision and Pattern Recognition (CVPR)}, 3498--3505. IEEE.

\bibitem[{Pf{\"u}lb, Gepperth, and Bagus(2021)}]{pfulb2021continual}
Pf{\"u}lb, B.; Gepperth, A.; and Bagus, B. 2021.
\newblock Continual learning with fully probabilistic models.
\newblock \emph{arXiv preprint arXiv:2104.09240}.

\bibitem[{Rolnick et~al.(2019)Rolnick, Ahuja, Schwarz, Lillicrap, and Wayne}]{rolnick2019experience}
Rolnick, D.; Ahuja, A.; Schwarz, J.; Lillicrap, T.; and Wayne, G. 2019.
\newblock Experience replay for continual learning.
\newblock \emph{Advances in Neural Information Processing Systems}, 32.

\bibitem[{Siahkamari et~al.(2020)Siahkamari, Xia, Saligrama, Casta{\~n}{\'o}n, and Kulis}]{siahkamari2020learning}
Siahkamari, A.; Xia, X.; Saligrama, V.; Casta{\~n}{\'o}n, D.; and Kulis, B. 2020.
\newblock Learning to approximate a Bregman divergence.
\newblock \emph{Advances in Neural Information Processing Systems}, 33: 3603--3612.

\bibitem[{Smith et~al.(2024)Smith, Hsu, Zhang, Hua, Kira, Shen, and Jin}]{smith2024continual}
Smith, J.~S.; Hsu, Y.-C.; Zhang, L.; Hua, T.; Kira, Z.; Shen, Y.; and Jin, H. 2024.
\newblock Continual Diffusion: Continual Customization of Text-to-Image Diffusion with C-LoRA.
\newblock \emph{Transactions on Machine Learning Research}.

\bibitem[{Smith et~al.(2023)Smith, Karlinsky, Gutta, Cascante-Bonilla, Kim, Arbelle, Panda, Feris, and Kira}]{smith2023coda}
Smith, J.~S.; Karlinsky, L.; Gutta, V.; Cascante-Bonilla, P.; Kim, D.; Arbelle, A.; Panda, R.; Feris, R.; and Kira, Z. 2023.
\newblock CODA-Prompt: COntinual Decomposed Attention-based Prompting for Rehearsal-Free Continual Learning.
\newblock In \emph{Proceedings of the IEEE/CVF Conference on Computer Vision and Pattern Recognition}, 11909--11919.

\bibitem[{Song, Meng, and Ermon(2020)}]{song2020denoising}
Song, J.; Meng, C.; and Ermon, S. 2020.
\newblock Denoising diffusion implicit models.
\newblock \emph{arXiv preprint arXiv:2010.02502}.

\bibitem[{Song et~al.(2021)Song, Sohl-Dickstein, Kingma, Kumar, Ermon, and Poole}]{song2021scorebased}
Song, Y.; Sohl-Dickstein, J.; Kingma, D.~P.; Kumar, A.; Ermon, S.; and Poole, B. 2021.
\newblock Score-Based Generative Modeling through Stochastic Differential Equations.
\newblock In \emph{International Conference on Learning Representations}.

\bibitem[{Sun et~al.(2024)Sun, Liang, Dong, Li, Ding, and Cong}]{sun2024create}
Sun, G.; Liang, W.; Dong, J.; Li, J.; Ding, Z.; and Cong, Y. 2024.
\newblock Create your world: Lifelong text-to-image diffusion.
\newblock \emph{IEEE Transactions on Pattern Analysis and Machine Intelligence}.

\bibitem[{Varshney et~al.(2021)Varshney, Verma, Srijith, Carin, and Rai}]{varshney2021cam}
Varshney, S.; Verma, V.~K.; Srijith, P.; Carin, L.; and Rai, P. 2021.
\newblock Cam-gan: Continual adaptation modules for generative adversarial networks.
\newblock \emph{Advances in Neural Information Processing Systems}, 34: 15175--15187.

\bibitem[{Vincent(2011)}]{vincent2011connection}
Vincent, P. 2011.
\newblock A connection between score matching and denoising autoencoders.
\newblock \emph{Neural Computation}, 23(7): 1661--1674.

\bibitem[{Wah et~al.(2011)Wah, Branson, Welinder, Perona, and Belongie}]{wah2011caltech}
Wah, C.; Branson, S.; Welinder, P.; Perona, P.; and Belongie, S. 2011.
\newblock {The Caltech-UCSD Birds-200-2011 Dataset}.
\newblock Technical Report CNS-TR-2011-001, California Institute of Technology.

\bibitem[{Wan et~al.(2025)Wan, Ren, Huang, Zhang, Deng, Bao, and Nie}]{wanunderstanding}
Wan, H.; Ren, S.; Huang, W.; Zhang, M.; Deng, X.; Bao, Y.; and Nie, L. 2025.
\newblock Understanding the Forgetting of (Replay-based) Continual Learning via Feature Learning: Angle Matters.
\newblock In \emph{International Conference on Machine Learning}.

\bibitem[{Wang et~al.(2024)Wang, Xie, Zhang, Huang, Su, and Zhu}]{wang2024hierarchical}
Wang, L.; Xie, J.; Zhang, X.; Huang, M.; Su, H.; and Zhu, J. 2024.
\newblock Hierarchical decomposition of prompt-based continual learning: Rethinking obscured sub-optimality.
\newblock \emph{Advances in Neural Information Processing Systems}, 36.

\bibitem[{Wang et~al.(2025)Wang, Li, Zhang, and Chen}]{wang2025cut}
Wang, X.; Li, S.-y.; Zhang, J.; and Chen, S. 2025.
\newblock Cut out and Replay: A Simple yet Versatile Strategy for Multi-Label Online Continual Learning.

\bibitem[{Wang, Huang, and Hong(2022)}]{wang2022s}
Wang, Y.; Huang, Z.; and Hong, X. 2022.
\newblock S-prompts learning with pre-trained transformers: An occam’s razor for domain incremental learning.
\newblock \emph{Advances in Neural Information Processing Systems}, 35: 5682--5695.

\bibitem[{Wang et~al.(2022{\natexlab{a}})Wang, Zhang, Ebrahimi, Sun, Zhang, Lee, Ren, Su, Perot, Dy et~al.}]{wang2022dualprompt}
Wang, Z.; Zhang, Z.; Ebrahimi, S.; Sun, R.; Zhang, H.; Lee, C.-Y.; Ren, X.; Su, G.; Perot, V.; Dy, J.; et~al. 2022{\natexlab{a}}.
\newblock Dualprompt: Complementary prompting for rehearsal-free continual learning.
\newblock In \emph{European Conference on Computer Vision}, 631--648. Springer.

\bibitem[{Wang et~al.(2022{\natexlab{b}})Wang, Zhang, Lee, Zhang, Sun, Ren, Su, Perot, Dy, and Pfister}]{wang2022learning}
Wang, Z.; Zhang, Z.; Lee, C.-Y.; Zhang, H.; Sun, R.; Ren, X.; Su, G.; Perot, V.; Dy, J.; and Pfister, T. 2022{\natexlab{b}}.
\newblock Learning to prompt for continual learning.
\newblock In \emph{Proceedings of the IEEE/CVF Conference on Computer Vision and Pattern Recognition}, 139--149.

\bibitem[{Yang et~al.(2025)Yang, Tian, Li, Zhang, Shen, Tong, and Wang}]{yang2025mmada}
Yang, L.; Tian, Y.; Li, B.; Zhang, X.; Shen, K.; Tong, Y.; and Wang, M. 2025.
\newblock Mmada: Multimodal large diffusion language models.
\newblock \emph{arXiv preprint arXiv:2505.15809}.

\bibitem[{Ye and Bors(2021{\natexlab{a}})}]{ye2021lifelong2}
Ye, F.; and Bors, A.~G. 2021{\natexlab{a}}.
\newblock Lifelong teacher-student network learning.
\newblock \emph{IEEE Transactions on Pattern Analysis and Machine Intelligence}, 44(10): 6280--6296.

\bibitem[{Ye and Bors(2021{\natexlab{b}})}]{ye2021lifelong}
Ye, F.; and Bors, A.~G. 2021{\natexlab{b}}.
\newblock Lifelong twin generative adversarial networks.
\newblock In \emph{IEEE International Conference on Image Processing}, 1289--1293. IEEE.

\bibitem[{Ye and Bors(2024)}]{ye2024online}
Ye, F.; and Bors, A.~G. 2024.
\newblock Online task-free continual generative and discriminative learning via dynamic cluster memory.
\newblock In \emph{Proceedings of the IEEE/CVF Conference on Computer Vision and Pattern Recognition}, 26202--26212.

\bibitem[{Yu et~al.(2020)Yu, Kumar, Gupta, Levine, Hausman, and Finn}]{yu2020gradient}
Yu, T.; Kumar, S.; Gupta, A.; Levine, S.; Hausman, K.; and Finn, C. 2020.
\newblock Gradient surgery for multi-task learning.
\newblock \emph{Advances in Neural Information Processing Systems}, 33: 5824--5836.

\bibitem[{Zaj{\k{a}}c et~al.(2023)Zaj{\k{a}}c, Deja, Kuzina, Tomczak, Trzci{\'n}ski, Shkurti, and Mi{\l}o{\'s}}]{zajkac2023exploring}
Zaj{\k{a}}c, M.; Deja, K.; Kuzina, A.; Tomczak, J.~M.; Trzci{\'n}ski, T.; Shkurti, F.; and Mi{\l}o{\'s}, P. 2023.
\newblock Exploring continual learning of diffusion models.
\newblock \emph{arXiv preprint arXiv:2303.15342}.

\bibitem[{Zenke, Poole, and Ganguli(2017)}]{zenke2017continual}
Zenke, F.; Poole, B.; and Ganguli, S. 2017.
\newblock Continual learning through synaptic intelligence.
\newblock In \emph{International Conference on Machine Learning}, 3987--3995. PMLR.

\bibitem[{Zhang et~al.(2024{\natexlab{a}})Zhang, Zhou, Lin, Ye, Zhu, Wang, Gao, Wang, and Liang}]{zhang2024clog}
Zhang, H.; Zhou, J.; Lin, H.; Ye, H.; Zhu, J.; Wang, Z.; Gao, L.; Wang, Y.; and Liang, Y. 2024{\natexlab{a}}.
\newblock CLoG: Benchmarking Continual Learning of Image Generation Models.
\newblock \emph{arXiv preprint arXiv:2406.04584}.

\bibitem[{Zhang et~al.(2024{\natexlab{b}})Zhang, Naseer, Chen, Shen, Khan, Zhang, and Khan}]{zhang2024s3a}
Zhang, S.; Naseer, M.; Chen, G.; Shen, Z.; Khan, S.; Zhang, K.; and Khan, F.~S. 2024{\natexlab{b}}.
\newblock S3a: Towards realistic zero-shot classification via self structural semantic alignment.
\newblock In \emph{Proceedings of the AAAI Conference on Artificial Intelligence}, volume~38, 7278--7286.

\bibitem[{Zhao et~al.(2024)Zhao, Sun, Wang, Suo, and Liu}]{zhao2024invertavatar}
Zhao, X.; Sun, J.; Wang, L.; Suo, J.; and Liu, Y. 2024.
\newblock InvertAvatar: Incremental GAN Inversion for Generalized Head Avatars.
\newblock In \emph{ACM SIGGRAPH 2024 Conference Papers}, 1--10.

\bibitem[{Zheng et~al.(2023)Zheng, Lu, Chen, and Zhu}]{zheng2023dpm}
Zheng, K.; Lu, C.; Chen, J.; and Zhu, J. 2023.
\newblock Dpm-solver-v3: Improved diffusion ode solver with empirical model statistics.
\newblock \emph{Advances in Neural Information Processing Systems}, 36: 55502--55542.

\end{thebibliography}

\clearpage
\appendix
\section{Preliminaries in SDE Diffusion Models} \label{sde_defination}
In this section, we introduce key preliminary concepts essential for the theoretical derivations that follow. Specifically, we discuss the framework of Stochastic Differential Equations (SDEs) and establish their equivalence with Denoising Diffusion Probabilistic Models (DDPMs), highlighting their shared formulation in modeling diffusion processes. This correspondence demonstrates that SDEs and DDPMs can be treated interchangeably in certain contexts.

\subsection{Fundamental Concepts}
Within the framework of stochastic differential equations (SDEs), controllable diffusion models are utilized to model the temporal evolution of data states under specified conditions or labels. Let \( x_t \) denote the data state at time \( t \). The forward SDE is defined as:
\begin{equation}
    dx_t = f(x_t, t) \, dt + g(t) \, dW_t,
\end{equation}
where \( f(x_t, t) \) represents the drift term, and \( g(t) \) signifies the diffusion coefficient. Here, \( W_t \) is a standard Wiener process (standard Brownian motion).

By incorporating the Fokker-Planck equation and reverse-time dynamics in \cite{anderson1982reverse,song2021scorebased}, the corresponding reverse SDE can be formulated with an additional drift adjustment:
\begin{equation}
    dx_t = \left( f(x_t, t) - g^2(t) \nabla_{x_t} \log p_t(x_t | y) \right) dt + g(t) \, dW_t,
\end{equation}
where the additional term \( \nabla_{x_t} \log p_t(x_t | y) \) corresponds to the gradient of the log-probability of the conditional distribution \( p_t(x_t | y) = p(x_t | y, t) \). Direct computation of this gradient is often infeasible for high-dimensional data and is thus typically approximated using a score estimator \( \epsilon_\theta \).

The score estimator \( \epsilon_\theta \) is employed to approximate the required \( \nabla_{x_t} \log p_t(x_t) \) for unconditional diffusion generation, and the optimization is achieved through the following loss:
\begin{align}
\epsilon_\theta^{*} = & \mathop{\arg\min}\limits_\theta \, \mathbb{E}_{t} \bigg\{ \lambda(t)\,\mathbb{E}_{p(x_0)} \big[
\mathbb{E}_{p(x_t|x_0)} \|\epsilon_\theta(x_t, t) \nonumber \\
& - \nabla_{x_t}\log p(x_t|x_0)\|_2^2 \big] \bigg\}
\label{optimization}
\end{align}
Here, $\lambda: [0, T] \in \mathbb{R}_{>0}$ is a positive weighting function, $t$ is uniformly sampled over $[0, T]$, $x_0 \sim p(x_0)$ and $x_t \sim p(x_t | x_0)$. With adequate data and sufficient model capacity, score matching ensures that the optimal solution to Eq.~\ref{optimization}, denoted by \( \epsilon_\theta^* \), equals \( \nabla_{x_t} \log p_t(x_t) \) for almost all \( x \) and \( t \).

To utilize the unconditional generation’s \( \epsilon_\theta^* \) to obtain the optimal parameters for conditional generation, we use \( p_t(x_t | y) = \frac{p_t(x_t, y)}{p_t(y)} \) to derive \( \nabla_{x_t} \log p_t(x_t | y) = \nabla_{x_t} \log p_t(x_t, y) - \nabla_{x_t} \log p_t(y) \). Since \( p_t(y) \) does not depend on \( x_t \), we have \( \nabla_{x_t} \log p_t(x_t | y) = \nabla_{x_t} \log p_t(x_t, y) \). Next, by the chain rule of differentiation, the gradient of the joint log-probability can be written as:
\begin{align}
\nabla_{x_t} \log p_t(x_t | y) 
&= \nabla_{x_t} \log p_t(x_t) + \nabla_{x_t} \log p_t(y | x_t) \nonumber \\
&= \epsilon_\theta^* + \nabla_{x_t} \log p_t(y | x_t)
\label{equ4}
\end{align}

\subsection{Equivalence Between DDPM and SDE Models} 
In the Denoising Diffusion Probabilistic Model (DDPM) framework \cite{ho2020denoising,nichol2021improved}, the data evolves through a forward process that progressively adds Gaussian noise to an initial sample \( x_0 \), resulting in a noisy sample \( x_t \). The probability distribution of \( x_t \) conditioned on \( x_0 \) can be written as:
\begin{equation}
    p_t(x_t | x_0) = \mathcal{N}(x_t; \bar{\alpha}_t x_0, \bar{\beta}_t^2 I),
\end{equation}
where \( \bar{\alpha}_t \) and \( \bar{\beta}_t \) are time-dependent coefficients that determine the scaling of the original data and the variance of the noise, respectively. Among them, \( \bar{\alpha}_t = \sqrt{\prod_{s=1}^t \alpha_s} \) and \( \bar{\beta}_t^2 = 1 - \bar{\alpha}_t^2 \).

From this, we can express the noisy sample \( x_t \) as:
\begin{equation}
    x_t = \bar{\alpha}_t x_0 + \bar{\beta}_t \varepsilon, \quad \varepsilon \sim \mathcal{N}(0, I).
\end{equation}

To recover \( x_0 \) from \( x_t \), rearrange the equation:
\begin{equation}
    x_0 = \frac{1}{\bar{\alpha}_t} x_t - \frac{\bar{\beta}_t}{\bar{\alpha}_t} \varepsilon.
\end{equation}

This equation elucidates how \( x_0 \) can be estimated from \( x_t \) using the scaling factor \( \bar{\alpha}_t \) and the noise variance \( \bar{\beta}_t \), both of which are determined by the forward process in DDPM.

In the SDE framework, the forward SDE for a Variance Preserving SDE (VP-SDE) is:
\begin{equation}
    dx = -\frac{1}{2} \beta(t) x \, dt + \sqrt{\beta(t)} \, dW,
\end{equation}
which has the solution:
\begin{equation}
    x_t = e^{-\frac{1}{2} \int_0^t \beta(s) ds} x_0 + \int_0^t \sqrt{\beta(s)} e^{-\frac{1}{2} \int_s^t \beta(u) du} dW_s.
\end{equation}

By choosing \( \bar{\alpha}(t) = e^{-\frac{1}{2} \int_0^t \beta(s) ds} \) and \( \bar{\sigma}^2(t) = 1 - \bar{\alpha}^2(t) \), we can write:
\begin{equation}
    x_t = \bar{\alpha}(t) x_0 + \bar{\sigma}(t) z, \quad z \sim \mathcal{N}(0, I).
\end{equation}

In this case, the relationship between \( x_0 \) and \( x_t \) is:
\begin{equation}
    x_0 = \frac{1}{\bar{\alpha}(t)} x_t - \frac{\bar{\sigma}(t)}{\bar{\alpha}(t)} z.
\end{equation}

This expression mirrors the form derived in the DDPM framework. To ensure consistency between the DDPM and SDE frameworks, we equate the coefficients from both equations. Thus, we have:
\begin{equation}
    \bar{\alpha}(t) = \bar{\alpha}_t \quad \text{and} \quad \bar{\sigma}(t) = \bar{\beta}_t.
\end{equation}

This shows that \( \bar{\sigma}(t) \) in the SDE framework can be expressed as \( \bar{\beta}_t \) from the DDPM framework when appropriate diffusion coefficients and scaling functions are selected.

\subsection{Score Approximation \& Noise Prediction} 
With the relationship between \( x_0 \) and \( x_t \) established, we now turn to the connection between the score function in SDEs and the noise prediction model in DDPMs. Certainly, this is merely a rough justification. A more comprehensive and rigorous proof of the equivalence between the two can be found in \cite{vincent2011connection}. In the SDE framework, the score function is defined as the gradient of the log-probability \( p_t(x_t | x_0) \) with respect to \( x_t \):
\begin{equation}
    \nabla_{x_t} \log p_t(x_t | x_0) = -\frac{1}{\bar{\sigma}(t)^2} (x_t - \bar{\alpha}(t) x_0).
\end{equation}

\begin{figure*}[htbp]
    \centering
    \includegraphics[width=1.0\textwidth]{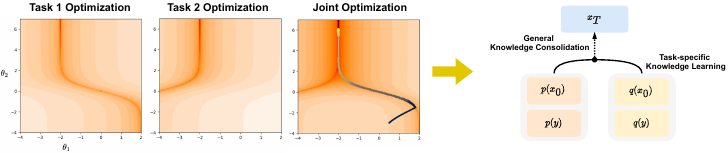}
    \caption{Borrowing from \cite{yu2020gradient}, during multi-task optimization, the gradients will eventually converge. Therefore, in a streaming scenario, there exists a portion of shared knowledge, which ensures that \( p_t(x_t | x_0, y) = q_t(x_t | x_0, y) \) and \( \nabla_{x_t} \mu(x_t, t) \approx \nabla_{x_t} \nu(x_t, t) \). This shared knowledge is crucial for the diffusion generator's ability to retain and transfer knowledge.}
    \label{appendix0}
    \vspace{-2ex}
\end{figure*}

Substituting \( \bar{\sigma}(t) = \bar{\beta}_t \) and \( \bar{\alpha}(t) = \bar{\alpha}_t \), we obtain:
\begin{equation}
    \nabla_{x_t} \log p_t(x_t | x_0) = -\frac{1}{\bar{\beta}_t^2} (x_t - \bar{\alpha}_t x_0).
\end{equation}

In DDPM, the model \( \varepsilon_\theta(x_t, t) \) is trained to predict the noise \( \varepsilon \), allowing us to express \( x_t = \bar{\alpha}_t x_0 + \bar{\beta}_t \varepsilon \).
\begin{equation}
    \nabla_{x_t} \log p_t(x_t | x_0) = -\frac{1}{\bar{\beta}_t} \varepsilon_\theta(x_t, t).
\end{equation}

With a large amount of \( x_0 \) training, \( \nabla_{x_t} \log p_t(x_t | x_0) \) will approximate \( \nabla_{x_t} \log p_t(x_t) = \epsilon_\theta(x_t, t) \). This establishes the connection between the score function in SDEs and the noise prediction model in DDPMs. Consequently, the score estimator \( \epsilon_\theta(x_t, t) \) in the SDE framework can be approximated by the noise predictor \( \varepsilon_\theta(x_t, t) \) from DDPMs.
\begin{equation} \label{equ6}
    \begin{aligned}
    \epsilon_\theta(x_t, t) &= \nabla_{x_t} \log p_t(x_t) \\
    &\approx \mathbb{E}_{p(x_0),\, p(x_t | x_0)} 
    \bigl[\nabla_{x_t} \log p_t(x_t | x_0)\bigr] \\
    &\approx -\frac{\varepsilon_\theta(x_t, t)}{\bar{\beta}_t}.
    \end{aligned}
\end{equation}

In summary, we have demonstrated that the relationship between \( x_0 \) and \( x_t \) in the SDE framework can be expressed similarly to that in the DDPM framework by appropriately selecting the diffusion coefficient \( \sigma(t) \) as \( \bar{\beta}_t \). Furthermore, the noise prediction model \( \varepsilon_\theta(x_t, t) \) in DDPM is approximately equivalent to the score estimator \( \epsilon_\theta(x_t, t) \) in the SDE framework, with a scaling factor \( \bar{\beta}_t \). This demonstrates the close correspondence between DDPMs and SDEs, and in the subsequent derivations, we will treat them as equivalent processes, having completed the detailed derivation.

\section{Cross-Task Diffusion Evolution} 
\label{ct_diffusion_loss}
From a Bayesian perspective, we analyze the forward transfer of diffusion models between two tasks \( \mathcal{T}_i \) and \( \mathcal{T}_j \). The visual space distributions of these tasks can be represented as \( p(x_0) \) and \( q(x_0) \), where \( x_0 \) encompasses all visual samples within \( \mathcal{T}_i \) and \( \mathcal{T}_j \). The prior spaces are expressed as \( p(y) \) and \( q(y) \), with \( y \) representing all labels within \( \mathcal{T}_i \) and \( \mathcal{T}_j \). As illustrated in Figure.~\ref{appendix0}, the relationship between these tasks can be visualized through the joint distribution of \( p(x_0, y) \) and \( q(x_0, y) \). The task-specific distributions \( p(x_0) \) and \( q(x_0) \) share certain overlapping regions in the visual space, signifying common knowledge that can be transferred between the tasks. The key challenge is to leverage these shared regions while accounting for task-specific differences.

We divide the training objective into two parts. The first part involves using \( \mathcal{T}_j \) data to enable the model to acquire task-specific knowledge pertinent to the target task. This is crucial for preventing underfitting on the target task and is referred to as the task-specific knowledge learning process. The second part focuses on retaining knowledge from the source task \( \mathcal{T}_i \). We refer to this process as the common knowledge consolidation process, aimed at preventing overfitting on the target domain and further strengthening the shared knowledge from the source task. This training strategy not only helps the model adapt better to new tasks but also ensures the effective retention of previously learned shared knowledge. In addition, based on the assumptions, we can establish the optimization relationship between the source and target tasks:
\begin{equation}
    \begin{aligned}
        \log q_t(x_t|y) &= \log \left( \int p_t(x_t|x_0, y) \frac{q_t(x_0|y)}{p_t(x_0|y)} p_t(x_0|y) \, dx_0 \right) \\
        &= \log \left( p_t(x_t|y) \mathbb{E}_{p_t(x_0|x_t, y)} \left[ \frac{q_t(x_0|y)}{p_t(x_0|y)} \right] \right) \\
        &= \log p_t(x_t|y) + \log \mathbb{E}_{p_t(x_0|x_t, y)} \left[ \frac{q_t(x_0|y)}{p_t(x_0|y)} \right],
    \end{aligned}
\end{equation}
where \( \mathbb{E}_{p_t(x_0|x_t, y)}[\cdot] \) denotes the conditional expectation under the posterior distribution \( p_t(x_0|x_t, y) \) given \( x_t \).

To refine this analysis, we compute the gradient of \( \log q_t(x_t|y) \) with respect to the data state \( x_t \) \footnote{\( x_t \) represents the shared knowledge in the \( p \) and \( q \), rather than diffused samples from a single distribution.}:
\begin{equation}
    \begin{aligned}
    \nabla_{x_t} \log q_t(x_t | y) 
    &= \nabla_{x_t} \log p_t(x_t | y) \\
    &+ \mathbb{E}_{p_t(x_0 | x_t, y)}
    \left[\nabla_{x_t} \log \frac{q_t(x_0 | y)}{p_t(x_0 | y)}\right].
    \end{aligned}
\end{equation}

Since \( \mathcal{T}_i \) and \( \mathcal{T}_j \) share some overlapping knowledge, we directly perform differential calculations on \( x_t \). The key quantity of interest is the term \( \nabla_{x_t} \log \mathbb{E}_{p_t(x_0|x_t, y)} \left[ \frac{q_t(x_0|y)}{p_t(x_0|y)} \right] \), which can be expressed as the difference in noise terms between the tasks. And combining with Eq.~\ref{equ4}, we obtain:
\begin{equation}
    \epsilon_\theta^{q} - \epsilon_\theta^{p} = \nabla_{x_t} \log \mathbb{E}_{p_t(x_0|x_t, y)} \left[ \frac{q_t(x_0|y)}{p_t(x_0|y)} \right] + \nabla_{x_t} \log \frac{p_t(y|x_t)}{q_t(y|x_t)},
\end{equation}
where \( \epsilon_\theta^{q} \) and \( \epsilon_\theta^{p} \) represent the noise approximations for tasks \( \mathcal{T}_j \) and \( \mathcal{T}_i \), respectively.

To achieve a simplified formulation, we employ Jensen's inequality, assuming equivalence between the conditional distributions of the original and synthesized imagery, i.e., \( p_t(x_0|x_t, y) = p_t(x_0|y) \) and \( q_t(x_0|x_t, y) = q_t(x_0|y) \). This alignment facilitates the derivation of a computationally amenable lower bound through the interchange of the logarithmic operation with the expectation:
\begin{equation}
    \begin{aligned}
    &\nabla_{x_t} \log \mathbb{E}_{p_t(x_0 | x_t, y)} 
    \left[\frac{q_t(x_0 | y)}{p_t(x_0 | y)}\right] \\
    &\geq \mathbb{E}_{p_t(x_0 | x_t, y)} 
    \left[\nabla_{x_t} \log \frac{q_t(x_0 | y)}{p_t(x_0 | y)}\right] \\
    &\approx -\nabla_{x_t} D_{KL}\Big(p_t(x_0 | x_t, y) \Big\| 
    q_t(x_0 | x_t, y)\Big).
    \end{aligned}
\end{equation}
where \( D_{KL} \) denotes the Kullback-Leibler (KL) divergence between the posteriors.

Finally, we re-express the gradient relationship as:
\begin{equation} \label{bound}
\begin{aligned}
&\epsilon_\theta^{q} - \epsilon_\theta^{p} 
- \nabla_{x_t} \log \mathbb{E}_{p_t(x_0 | x_t, y)} 
\left[\frac{q_t(x_0 | y)}{p_t(x_0 | y)}\right] \\
&\quad + \nabla_{x_t} \log \frac{q_t(y | x_t)}{p_t(y | x_t)} \\
\leq\ &\epsilon_\theta^{q} - \epsilon_\theta^{p} 
+ \nabla_{x_t} D_{KL}\Big(p_t(x_0 | x_t, y) 
\parallel q_t(x_0 | x_t, y)\Big) \\
&\quad + \nabla_{x_t} \log \frac{q_t(y | x_t)}{p_t(y | x_t)}.
\end{aligned}
\end{equation}

Thus, the evolution of cross-task diffusion can be systematically characterized through the gradient of the KL divergence, offering a principled framework to govern the conditional diffusion process across tasks. This framework not only ensures the fidelity of the generated samples to their respective tasks but also captures the intrinsic inter-task relationship in a mathematically coherent manner.

Applying Bayes' theorem to decompose \( p_t(x_0|x_t, y) \), we obtain \( p_t(x_0|x_t, y) = \frac{p_t(x_0|x_t)p_t(y|x_0)}{p_t(y|x_t)} \). By performing an analogous decomposition for \( q_t(x_0|x_t, y) \), we arrive at:

\begin{equation} \label{cross_sde}
    \begin{aligned}
    &\nabla_{x_t} D_{KL}\Big(p_t(x_0 | x_t, y) \parallel q_t(x_0 | x_t, y)\Big) \\
    &= \nabla_{x_t} \int p_t(x_0 | x_t, y) 
    \log \frac{\frac{p_t(x_0 | x_t)p_t(y | x_0)}{p_t(y | x_t)}}{\frac{q_t(x_0 | x_t)q_t(y | x_0)}{q_t(y | x_t)}} dx_0 \\
    &= \nabla_{x_t} \int p_t(x_0 | x_t, y) 
    \Bigg[\log \frac{p_t(x_0 | x_t)}{q_t(x_0 | x_t)} 
    + \log \frac{p_t(y | x_0)}{q_t(y | x_0)} \\
    &\quad + \log \frac{q_t(y | x_t)}{p_t(y | x_t)}\Bigg] dx_0 \\
    &= \nabla_{x_t} \int p_t(x_0 | x_t, y) 
    \log \frac{p_t(x_0 | x_t)}{q_t(x_0 | x_t)} dx_0 \\
    &\quad + \nabla_{x_t} \int p_t(x_0 | x_t, y) 
    \log \frac{p_t(y | x_0)}{q_t(y | x_0)} dx_0 \\
    &\quad + \nabla_{x_t} \log \frac{q_t(y | x_t)}{p_t(y | x_t)} 
    \int p_t(x_0 | x_t, y) dx_0 \\
    &= \underbrace{\nabla_{x_t} \int \frac{p_t(x_0 | x_t)p_t(y | x_0)}{p_t(y | x_t)} 
    \log \frac{p_t(x_0 | x_t)}{q_t(x_0 | x_t)} dx_0}_{\text{Unconditional Knowledge Consistency}} \\
    &\quad + \underbrace{\nabla_{x_t} \int \frac{p_t(x_0 | x_t)p_t(y | x_0)}{p_t(y | x_t)} 
    \log \frac{p_t(y | x_0)}{q_t(y | x_0)} dx_0}_{\text{Prior Knowledge Consistency}} \\
    &\quad + \underbrace{\nabla_{x_t} \log \frac{q_t(y | x_t)}{p_t(y | x_t)}}_{\text{Simplifiable Aspect}}.
    \end{aligned}
\end{equation}

For the unconditional knowledge consistency term, we assume that, aside from the clean sample \(x_0\) itself, the correlation between \(y\) and most diffused samples \(x_t\) diminishes as the diffusion process adds noise. Furthermore, we intentionally avoid introducing an explicit classifier that couples \(x_t\) with \(y\). Under this formulation, \(p_t(y| x_t)\) can be regarded as irrelevant to the differentiation variable \(x_t\), and \(p_t(y| x_0)\) can be reformulated as an external expectation term. Consequently, the unconditional knowledge consistency term simplifies to:
\begin{equation}
    \begin{aligned}
    \nabla_{x_t}\!\int 
    \frac{p_t(x_0| x_t)\,p_t(y| x_0)}{p_t(y| x_t)}
    \log\frac{p_t(x_0| x_t)}{q_t(x_0| x_t)}\,dx_0 \\[4pt]
    = \nabla_{x_t}\!\int 
    p_t(x_0| x_t)\,
    \log\frac{p_t(x_0| x_t)}{q_t(x_0| x_t)}\,dx_0 \\[4pt]
    = \nabla_{x_t} D_{\mathrm{KL}}\Bigl(p_t(x_0| x_t)\,\big\|\,q_t(x_0| x_t)\Bigr).
    \end{aligned}
\end{equation}

Hence, only the gradient of the KL divergence between the two posteriors \(p_t(x_0| x_t)\) and \(q_t(x_0| x_t)\) remains relevant to optimization, while the label‑related factors play no role.

In the context of unconditional diffusion generation, both the generative distribution \( p(x_0|x_t) \) and the approximate posterior \( q(x_0|x_t) \) are typically modeled as Gaussian distributions of the form \( \mathcal{N}(x_0; \mu(x_t, t), \sigma_t^2 I) \) and \( \mathcal{N}(x_0; \nu(x_t, t), \sigma_t^2 I) \), respectively. The Kullback–Leibler (KL) divergence between these two Gaussians, which share the same isotropic variance \( \sigma_t^2 I \), reduces to a simple expression involving their means, namely \( D_{KL} = \frac{(\mu(x_t, t) - \nu(x_t, t))^2}{2\sigma_t^2} \). 

Assuming that both \( \mu(x_t, t) \) and \( \nu(x_t, t) \) are differentiable functions of \( x_t \), and that their dependence on \( x_t \) is continuous, we can take the gradient of the KL divergence with respect to \( x_t \). This yields:
\begin{equation}
    \begin{aligned}
    \nabla_{x_t} D_{KL} 
    &= \frac{(\mu(x_t, t) - \nu(x_t, t))}{\sigma_t^2} 
    \big( \nabla_{x_t} \mu(x_t, t) - \nabla_{x_t} \nu(x_t, t) \big) \\
    &\approx \delta \cdot \frac{\mu(x_t, t) - \nu(x_t, t)}{\sigma_t^2}.
    \end{aligned}
\end{equation}
where \( \delta \) represents a small perturbation determined based on the shared knowledge observed in Figure.~\ref{appendix0}.

Consequently, the term \( \nabla_{x_t} D_{\text{KL}}(p_t(x_0|x_t) \| q_t(x_0|x_t)) \) can be rephrased as:
\begin{equation}
    \begin{aligned}
    &\nabla_{x_t} D_{KL}\Big(p_t(x_0 | x_t) \parallel q_t(x_0 | x_t)\Big) \\
    &= \frac{\bar{\alpha}_t^2}{\bar{\beta}_t^2} \times \Big[\mu_\theta(x_t, t) - \nu_\theta(x_t, t)\Big].
    \end{aligned}
\end{equation}
where \( x_0 = \frac{1}{\bar{\alpha}_t} x_t - \frac{\bar{\beta}_t}{\bar{\alpha}_t} \varepsilon \), implying that \( \sigma_t^2 = \frac{\bar{\beta}_t^2}{\bar{\alpha}_t^2} \). The parameter \( \theta \) represents the network parameters utilized in the reparameterization technique to fit the mean. \( \varepsilon \) is standard normal noise.

Similarly, for the term representing Prior Knowledge Consistency, the core focus is on directly describing the differences in the conditional distributions of labels \( y \) given \( x_0 \). Here, the labels play a direct comparison role, and the label-related probability distributions directly impact the optimization objective. In contrast, \( p_t(y|x_t) \) does not affect the optimization objective related to \( x_0 \) and can be neglected to some extent. Therefore, this term can be transformed into:
\begin{equation}
    \begin{aligned}
    &\nabla_{x_t} \int \frac{p_t(x_0 | x_t)p_t(y | x_0)}{p_t(y | x_t)} 
    \log \frac{p_t(y | x_0)}{q_t(y | x_0)} dx_0 \\
    &\propto \nabla_{x_t} \int p_t(x_0 | x_t)p_t(y | x_0) 
    \log \frac{p_t(y | x_0)}{q_t(y | x_0)} dx_0 \\
    &= \int \nabla_{x_t} p_t(x_0 | x_t) p_t(y | x_0) 
    \log \frac{p_t(y | x_0)}{q_t(y | x_0)} dx_0 \\
    &= \mathbb{E}_{\nabla_{x_t} p(x_0 | x_t)} 
    \Big[D_{KL}\big(p_t(y | x_0) \parallel q_t(y | x_0)\big)\Big].
    \end{aligned}
\end{equation}

In diffusion models, \( p_t(x_0|x_t) \) is typically modeled as a Gaussian distribution \( \mathcal{N}(x_0^p; \mu(x_t, t), \sigma_t^2 I) \). Thus, the probability density function of \( p_t(x_0|x_t) \) is given by:
\begin{equation}
    \begin{aligned}
    p_t(x_0 | x_t) &= \frac{1}{\sqrt{(2\pi \sigma_t^2)^d}}
    \times \exp\left(-\frac{1}{2\sigma_t^2} \| x_0^p - \mu(x_t, t) \|^2 \right) \\
    &\quad \text{where } x_0^p \in \mathcal{T}_i.
    \end{aligned}
\end{equation}

When taking the derivative with respect to \( x_t \), only \( \mu(x_t, t) \) depends on \( x_t \), hence:
\begin{equation}
    \begin{aligned}
        \nabla_{x_t} \log p_t(x_0|x_t) 
        &= \nabla_{x_t} \left( -\frac{1}{2\sigma_t^2} \| x_0^p - \mu(x_t, t) \|^2 \right) \\
        &= \frac{1}{\sigma_t^2} (x_0^p - \mu(x_t, t)) \nabla_{x_t} \mu(x_t, t).
    \end{aligned}
\end{equation}

Consequently, \( \nabla_{x_t} p_t(x_0|x_t) \) can be written as:
\begin{equation}
    \begin{aligned}
        \nabla_{x_t} p_t(x_0|x_t) &= p_t(x_0|x_t) \frac{1}{\sigma_t^2} (x_0^p - \mu(x_t, t)) \nabla_{x_t} \mu(x_t, t) \\
        &= \frac{\bar{\alpha}_t^2}{\bar{\beta}_t^2} p_t(x_0|x_t) (x_0^p - \mu(x_t, t)) \nabla_{x_t} \mu(x_t, t).
    \end{aligned}
\end{equation}

We now substitute the gradient \( \nabla_{x_t} p_t(x_0|x_t) \) into \( \mathbb{E}_{\nabla_{x_t} p_t(x_0|x_t)} \left[ D_{KL}(p_t(y|x_0) \| q_t(y|x_0)) \right] \):
\begin{equation}
    \begin{aligned}
    &\mathbb{E}_{\nabla_{x_t} p_t(x_0 | x_t)} 
    \Big[D_{KL}\big(p_t(y | x_0) \parallel q_t(y | x_0)\big)\Big] \\
    &= \int \nabla_{x_t} p_t(x_0 | x_t) p_t(y | x_0) 
    \log \frac{p_t(y | x_0)}{q_t(y | x_0)} dx_0 \\
    &= \frac{\bar{\alpha}_t^2}{\bar{\beta}_t^2} 
    \int p_t(x_0 | x_t) (x_0^p - \mu(x_t, t)) \nabla_{x_t} \mu(x_t, t) p_t(y | x_0) \\
    &\quad \times \log \frac{p_t(y | x_0)}{q_t(y | x_0)} dx_0.
    \end{aligned}
\end{equation}

Given that \( p_t(x_0 | x_t) \) can be expressed as \( \mathcal{N}(x_0^p; \mu(x_t, t), \sigma_t^2 I) \), and \( x_0^p = \frac{1}{\bar{\alpha}_t} x_t^p - \frac{\bar{\beta}_t}{\bar{\alpha}_t} \varepsilon \), then \( \mu(x_t, t) \) can also be represented as \( \frac{1}{\bar{\alpha}_t} x_t^p - \frac{\bar{\beta}_t}{\bar{\alpha}_t} \varepsilon_\theta(x_t^p, t) \). Consequently, the difference \( x_0^p - \mu(x_t, t) \) becomes \( \frac{\bar{\beta}_t}{\bar{\alpha}_t} \left[\varepsilon_\theta(x_t^p, t) - \varepsilon\right] \), where \( \varepsilon_\theta(x_t^p, t) \) denotes the noise prediction output of the unconditional diffusion model at time \( t \) in task \( \mathcal{T}_i \). Similarly, the gradient \( \nabla_{x_t} \mu(x_t, t) \) is given by \( \frac{1}{\bar{\alpha}_t} I - \frac{\bar{\beta}_t}{\bar{\alpha}_t} \nabla_{x_t} \varepsilon_\theta(x_t^p, t) \), where \( \nabla_{x_t} \varepsilon_\theta(x_t^p, t) \) is the Jacobian of the noise prediction model in \( \mathcal{T}_i \). By optimizing \( \varepsilon_\theta(x_t^p, t) \) to approximate \( \varepsilon \), we implicitly optimize \( \mu(x_t, t) \) and its gradient without explicitly computing \( \nabla_{x_t} \mu(x_t, t) \). As a result, the term \( (x_0^p - \mu(x_t, t)) \nabla_{x_t} \mu(x_t, t) \) simplifies to:
\begin{equation} \label{equ23}
    \begin{aligned}
    (x_0^p - \mu(x_t, t)) \nabla_{x_t} \mu(x_t, t)
    &\propto \frac{\bar{\beta}_t}{\bar{\alpha}_t} \|\varepsilon_\theta(x_t^p, t) - \varepsilon\|_2^2 \\
    &= \frac{\bar{\beta}_t}{\bar{\alpha}_t} \|\beta_t \epsilon_\theta^p + \varepsilon\|_2^2.
    \end{aligned}
\end{equation}
where this step is based on the approximations derived in Eq.~\ref{equ6} and Eq.~\ref{equ23}.

Ultimately, the term for Prior Knowledge Consistency can be transformed into a form that is optimizable for SDE diffusion models:
\begin{equation}
    \begin{aligned}
    &\nabla_{x_t} \int \frac{p_t(x_0 | x_t)p_t(y | x_0)}{p_t(y | x_t)} 
    \log \frac{p_t(y | x_0)}{q_t(y | x_0)} dx_0 \\
    &\propto \frac{\bar{\alpha}_t}{\bar{\beta}_t} 
    \|\beta_t \epsilon_\theta^p + \varepsilon\|_2^2 \mathbb{E}_{p_t(x_0 | x_t)} 
    D_{KL}\big(p_t(y | x_0) \parallel q_t(y | x_0)\big), \\
    &\propto \frac{\bar{\alpha}_t}{\bar{\beta}_t} \mathbb{E}_{p_t(x_0 | x_t)} 
    D_{KL}\big(p_t(y | x_0) \parallel q_t(y | x_0)\big),
    \end{aligned}
\end{equation}
Here, the term \(\|\beta_t \epsilon_\theta^p + \varepsilon\|_2^2\) involves the model parameters \(\epsilon_\theta^{p}\) optimized from the previous task and the standard normal distribution \(\varepsilon\), making it non-optimizable. Therefore, we consider this term as a learnable scaling factor that the network can adjust on its own.

For Simplifiable Aspect term, in the context of cross-task knowledge retention via diffusion models, \( x_t \) represents the shared knowledge between tasks \( \mathcal{T}_i \) and \( \mathcal{T}_j \). As the noise accumulates during diffusion, the correlation between \( x_t \) and the label \( y \) diminishes, reflecting the increasing abstraction of task-specific information. In this setting, the term \( \nabla_{x_t} \log \frac{q_t(y | x_t)}{p_t(y | x_t)} \), which quantifies the gradient of the divergence between label-conditioned distributions across tasks, becomes increasingly insignificant. Mathematically, as \( x_t \) evolves toward a noisy state, its dependency on \( y \) weakens, and the conditional distributions of \( y \) given \( x_t \) from both tasks become approximately equal. Consequently, the gradient of their ratio \( \log \frac{q_t(y | x_t)}{p_t(y | x_t)} \) tends towards zero, justifying its omission from the objective function. This simplification streamlines the optimization process, allowing the model to focus on the shared knowledge represented by \( x_t \), while reducing the computational cost associated with task-specific label divergence and label regressor parameters, thereby improving efficiency.

As shown above, the upper bound in Eq.~\ref{bound} can be expressed as:
\begin{equation}
    \begin{aligned}
        \mathcal{L}_{UB} &= \kappa (\epsilon_\theta^{q} - \epsilon_\theta^{p}) + \lambda\,\mathcal{L}_{UKC} + \eta\,\mathcal{L}_{PKC} \\
        &= \kappa \mathcal{L}_{IKC} + \lambda\,\mathcal{L}_{UKC} + \eta\,\mathcal{L}_{PKC}.
    \end{aligned}
\end{equation}
\begin{equation}
    \left\{
    \begin{aligned}
        &\mathcal{L}_{IKC} = \epsilon_\theta^{q} - \epsilon_\theta^{p}, \\
        &\mathcal{L}_{UKC} = \frac{\bar{\alpha}_t^2}{\bar{\beta}_t^2} \left[ \mu_\theta(x_t, t) - \nu_\theta(x_t, t) \right], \\
        &\mathcal{L}_{PKC} = \frac{\bar{\alpha}_t}{\bar{\beta}_t} \mathbb{E}_{p_t(x_0|x_t)} D_{KL}(p_t(y|x_0) || q_t(y|x_0)),
    \end{aligned}
    \right.
\end{equation}
where \(\kappa\), \(\lambda\), and \(\eta\) are weighting hyperparameters that balance the contributions of three knowledge consistency components in the total upper bound loss \(\mathcal{L}_{UB}\). 

In summary, through a detailed analysis of cross-task knowledge retention, we have developed a robust optimization strategy for shared knowledge, enabling the seamless adaptation of diffusion models across multiple tasks in continual learning scenarios. The derived loss functions, encapsulating the critical components of knowledge consistency, provide a principled approach to balancing the retention of prior task knowledge while accommodating the nuances of new tasks. This work not only furthers our understanding of diffusion models in a multi-task context but also lays the foundation for more efficient and scalable generative models capable of leveraging the inherent relationships between tasks in a dynamic, continual learning setup.

\begin{figure*}[h]
    \centering
    \includegraphics[scale=0.65]{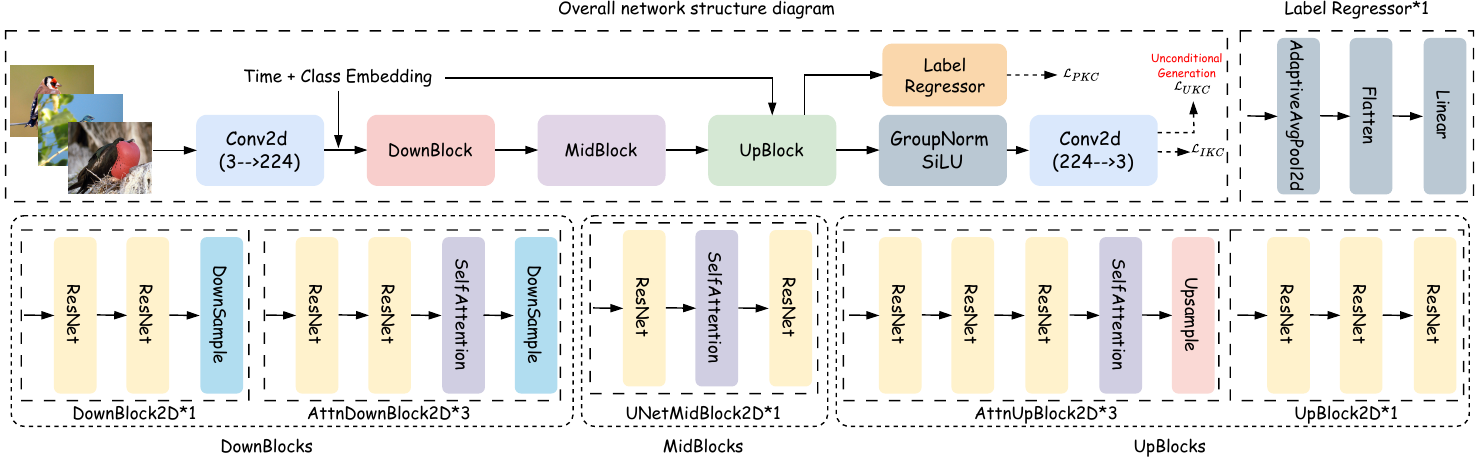}
    \caption{Overall architecture of the CCD framework. The backbone follows a U-Net-style encoder–decoder structure with residual and attention units, and is augmented with additional heads to support intermediate knowledge consistency objectives.}
  \label{frameworks}
\end{figure*}

\section{Model Architecture Diagram} \label{model_architecture}
Figure~\ref{frameworks} presents a detailed schematic of the diffusion backbone used in all of our experiments.  The design follows the ``U‑Net with cross‑task hooks'' blueprint popularised in contemporaneous diffusion work, but is augmented with three novel pathways that are required by the Continual Consistency Diffusion (CCD) training objectives.

\section{Method-specific Implementation Details}

\setlist[description]{
    leftmargin=2.5em,      
    labelindent=1em,       
    style=nextline,        
    itemsep=0.2em,         
    parsep=0pt             
}

\subsection{Baseline Method Implementations}
\textbf{ER:} Implements a FIFO-based buffer to store past samples, where the replay batch is combined with the current task batch to match the training batch size used in non-buffer-based methods, ensuring strict consistency. During each update, replayed samples are concatenated with the current batch for joint training.
\begin{description}
    \item[Buffer Size (512 or 2560 or 5120):] Controls the number of stored past samples. Larger buffers reduce GCF but increase memory usage.
    \item[Replay Batch Size (100):] Number of samples drawn from memory per step.
\end{description}

\noindent\textbf{L2 Regularization:} Prevents GCF by constraining current model parameters to remain close to those from the previous task. The method maintains a frozen teacher model from the previous task and applies L2 penalty on parameter deviations: $\mathcal{L}_{L2} = \sum_{i} \|\theta_i - \theta_{teacher,i}\|_2^2$ with weight $\lambda_{L2} = 50.0$. This direct parameter constraint helps preserve previous task knowledge while allowing adaptation to new tasks.

\noindent\textbf{A-GEM:} Uses gradient projection with an episodic memory to prevent interference with previous tasks. It computes a reference gradient on replay data and projects the current gradient to ensure a non-negative dot product with the reference.
\begin{description}
    \item[Buffer Size (512 or 2560 or 5120):] Representative samples from prior tasks; more memory improves gradient accuracy.
    \item[Gradient Projection Rule:] Ensures updates do not decrease past-task performance, balancing stability and plasticity.
\end{description}

\noindent\textbf{DCM:} Adopts a hierarchical memory structure that organizes stored samples into adaptive clusters for diverse and efficient replay. It dynamically creates, merges, or updates clusters based on knowledge discrepancy measures.
\begin{description}
    \item[Buffer Size (512 or 2560 or 5120):] Total number of stored samples across all clusters, defining overall replay capacity.
    \item[Cluster Capacity (64 or 128 or 256):] Maximum samples per cluster; when exceeded, the most redundant sample is removed.
    \item[Expansion Threshold (1500 or 2000 or 2500):] A new cluster is created if a sample is farther than this threshold from all current prototypes.
    \item[Maximum Clusters (20):] Upper bound on the total number of clusters; exceeding this triggers merging of the two most similar clusters.
    \item[Prototype Selection (Square\_Error):] Each cluster maintains a prototype minimizing intra-cluster distances, updated periodically to reduce computation cost.
\end{description}

\noindent\textbf{EWC:} Estimates parameter importance via the Fisher Information Matrix; a quadratic penalty (diagonal approximation) constrains critical parameter drift.
\begin{description}
    \item[Regularization Weight $\lambda_{EWC} = 5.0$:] Balances old-task retention and new-task learning; higher values increase stability but reduce adaptability.
    \item[Fisher Diagonal:] Measures parameter sensitivity to past tasks, guiding which weights are most protected.
\end{description}

\noindent\textbf{MAS:} Measures parameter importance via the L2 norm of model outputs with respect to each parameter and accumulates importance across training.
\begin{description}
    \item[Regularization Weight $\lambda_{MAS} = 5e-5$:] Higher values preserve past knowledge more aggressively but may restrict new learning.
\end{description}

\noindent\textbf{SI: } Tracks parameter trajectories to compute importance scores, which are used to regularize updates after each task.
\begin{description}
    \item[Regularization Weight $\lambda_{SI} = 5.0$:] Controls the trade-off between stability and plasticity.
    \item[Stability Factor $\epsilon_{SI} = 0.01$:] Prevents division by near-zero parameter changes, ensuring numerical stability.
\end{description}

\noindent\textbf{LwF:} Also known as Knowledge Distillation (KD), this method preserves knowledge from previous tasks by aligning the current model’s outputs with those of a frozen teacher from the prior task, using an MSE loss on noise predictions weighted by $\lambda_{LwF}=0.01$.

\noindent\textbf{C-LoRA:} A parameter-efficient continual learning approach that combines full fine-tuning for the first task with Low-Rank Adaptation for subsequent tasks. The method employs the following strategy:
\begin{description}
    \item[Task 0:] Full fine-tuning of the entire UNet model to establish a strong backbone.
    \item[Subsequent Tasks:] Inject LoRA adapters with rank $r=8$, $\alpha=8$, dropout=0.1, targeting modules \texttt{['to\_k', 'to\_q', 'to\_v', 'to\_out.0', 'conv1', 'conv2']}.
    \item [Selective Training:] Only LoRA parameters and class embeddings are trainable; backbone remains frozen.
    \item [Memory Management:] Past class embeddings are preserved to prevent weight decay during training.
    \item [Task-specific Inference:] For each task, the model reconstructs the appropriate architecture by loading the backbone and injecting task-specific LoRA weights.
\end{description}

\subsection{CCD Implementation Details}

Our Continual Consistency Diffusion implements three consistency losses with specific computational strategies:

\textbf{Inter-task Knowledge Consistency ($\mathcal{L}_{IKC}$):} Employs Bregman divergence with Fisher Information-based preconditioner. The Fisher diagonal is computed as the squared mean of teacher model outputs, providing curvature-aware distance metrics between teacher and student score functions.

\textbf{Unconditional Knowledge Consistency ($\mathcal{L}_{UKC}$):} Applies time-dependent masking with threshold $t < 700$ to emphasize semantically critical mid-diffusion phases. The reverse-time mean functions are computed using the reparameterization $\mu(x_t, t) = \frac{1}{\sqrt{\alpha_t}}x_t - \frac{1-\alpha_t}{\sqrt{\alpha_t(1-\bar{\alpha}_t)}}\epsilon_\theta(x_t, t)$.

\textbf{Prior Knowledge Consistency ($\mathcal{L}_{PKC}$):} Utilizes KL divergence between mid-diffusion features at timestep $t=0$. The loss enforces semantic consistency by comparing prior embeddings from current and past models on replay samples.

\subsection{Hyperparameter Configuration}
Table~\ref{tab:implementation_params} presents the hyperparameters used across all methods, extracted from our experimental codebase.

\begin{table*}[h]
    \centering
    \small
    \renewcommand{\arraystretch}{1.1}
    \setlength{\tabcolsep}{2.5mm}
    \begin{tabular}{lcc}
        \toprule
        \textbf{Parameter} & \textbf{Value} & \textbf{Description} \\
        \midrule
        \multicolumn{3}{c}{\textbf{CCD Framework}} \\
        \midrule
        $\kappa$ (IKC weight) & $1 \times 10^{-5} \sim 1 \times 10^{-7}$ & Inter-task consistency \\
        $\lambda$ (UKC weight) & $1 \times 10^{-5} \sim 1 \times 10^{-7}$ & Unconditional consistency \\
        $\eta$ (PKC weight) & $1 \times 10^{-5} \sim 1 \times 10^{-7}$ & Prior knowledge consistency \\
        \midrule
        \multicolumn{3}{c}{\textbf{Regularization Methods}} \\
        \midrule
        EWC weight ($\lambda_{EWC}$) & 5.0 & Fisher penalty coefficient \\
        L2 weight ($\lambda_{L2}$) & 50.0 & Parameter regularization \\
        MAS weight ($\lambda_{MAS}$) & $5 \times 10^{-5}$ & Importance-weighted penalty \\
        SI weight ($\lambda_{SI}$) & 5.0 & Synaptic importance penalty \\
        SI epsilon ($\epsilon_{SI}$) & 0.01 & Numerical stability term \\
        LwF weight ($\lambda_{LwF}$) & 0.01 & Knowledge distillation penalty \\
        \midrule
        \multicolumn{3}{c}{\textbf{DCM Configuration}} \\
        \midrule
        Cluster capacity & 64/128/256 & Max samples per cluster \\
        Expansion threshold & 1500/2000/2500 & New cluster creation threshold \\
        Maximum clusters & 20 & Upper bound on cluster count \\
        KDM type & Square\_Error & Knowledge discrepancy measure \\
        \midrule
        \multicolumn{3}{c}{\textbf{C-LoRA Configuration}} \\
        \midrule
        LoRA rank ($r$) & 8 & Low-rank approximation rank \\
        LoRA alpha ($\alpha$) & 8 & Scaling parameter \\
        LoRA dropout & 0.1 & Dropout rate for LoRA layers \\
        \midrule
        \multicolumn{3}{c}{\textbf{Memory \& Training Setup}} \\
        \midrule
        Buffer sizes & 512/2560/5120 & Rehearsal memory capacity \\
        Replay batch size & 100 & Memory sampling size \\
        Training batch size & 200 & Total training batch size \\
        Diffusion timesteps & 1000 & Forward process steps \\
        Inference steps & 50 & DDIM sampling steps \\
        Learning rate & $1 \times 10^{-3}$ & Adam optimizer \\
        \bottomrule
    \end{tabular}
    \caption{Complete hyperparameter configuration for all implemented methods.}
    \label{tab:implementation_params}
\end{table*}


\section{Perceptual Metrics}
In this section, we select three representative benchmarks to examine changes in perceptual loss, specifically the LPIPS metric. For LPIPS, we adapt the FID computation procedure to the CDG pipeline and compare against the most relevant baseline, ER. Detailed results are shown in Table.~\ref{tab:per_metrics_trimmed}.

From the results, we observe that LPIPS shows limited variation under our CCD framework compared to ER. This is largely attributed to inherent limitations of the UNet architecture and the pixel resolution used during training and evaluation. Notably, the baseline ER method already performs poorly on OxfordPets-5T and Flowers102-10T, posing additional challenges for CCD optimization. This highlights a key direction for future work, developing more capable model architectures to further improve perceptual quality.

\begin{table}[t]
    \centering
    \small
    \renewcommand{\arraystretch}{1.0}
    \setlength{\tabcolsep}{1.2mm}
    \begin{tabular}{@{}c cc cc cc@{}}
        \toprule
        \rowcolor{gray!15}
        \textbf{Method}
        & \multicolumn{2}{c}{\textbf{MNIST-5T}} 
        & \multicolumn{2}{c}{\textbf{OxfordPets-5T}}
        & \multicolumn{2}{c}{\textbf{Flowers102-10T}} \\
        \cmidrule(lr){2-3} \cmidrule(lr){4-5} \cmidrule(lr){6-7}
        & MP↓ & IMP↓ & MP↓ & IMP↓ & MP↓ & IMP↓ \\
        \midrule
        \rowcolor{gray!10} \multicolumn{7}{c}{\textbf{Storage Rehearsal Methods (512 buffer)}} \\
        ER 
        & 0.38 & 0.54 & 0.79 & 0.78 & 0.65 & 0.68 \\
        \rowcolor{cyan!15}
        CCD + ER 
        & 0.39 & 0.54 & 0.78 & 0.77 & 0.65 & 0.67 \\
        \midrule
        \rowcolor{gray!10} \multicolumn{7}{c}{\textbf{Storage Rehearsal Methods (2560 buffer)}} \\
        ER 
        & 0.37 & 0.43 & 0.78 & 0.78 & 0.63 & 0.66 \\
        \rowcolor{cyan!15}
        CCD + ER 
        & 0.37 & 0.43 & 0.78 & 0.77 & 0.65 & 0.66 \\
        \midrule
        \rowcolor{gray!10} \multicolumn{7}{c}{\textbf{Storage Rehearsal Methods (5120 buffer)}} \\
        ER 
        & 0.37 & 0.42 & 0.78 & 0.77 & 0.64 & 0.65 \\
        \rowcolor{cyan!15}
        CCD + ER 
        & 0.37 & 0.42 & 0.78 & 0.77 & 0.64 & 0.65 \\
        \bottomrule
    \end{tabular}
    \vspace{-1ex}
    \caption{LPIPS comparison across datasets and buffer sizes.}
    \vspace{-4ex}
    \label{tab:per_metrics_trimmed}
\end{table}

\section{Visualizations} \label{visualizations}
In this section, we present the visualizations of samples generated for the first task across all datasets at the final training stage. All results are sampled based on class labels and produced using a model trained with a buffer size of 5120, ensuring a fair and unbiased comparison.

\begin{figure}[h]
    \centering
    \begin{subfigure}{0.48\textwidth}
        \centering
        \includegraphics[width=\textwidth]{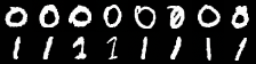}
        \caption{The Original Results.}
    \end{subfigure}
    \hfill
    \begin{subfigure}{0.48\textwidth}
        \centering
        \includegraphics[width=\textwidth]{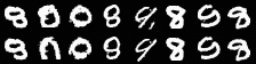}
        \caption{Our CCD Results.}
    \end{subfigure}
    \hfill
    \begin{subfigure}{0.48\textwidth}
        \centering
        \includegraphics[width=\textwidth]{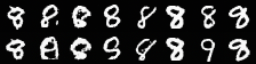}
        \caption{ER Results.}
    \end{subfigure}
    \begin{subfigure}{0.48\textwidth}
        \centering
        \includegraphics[width=\textwidth]{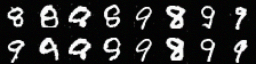}
        \caption{A-GEM Results.}
    \end{subfigure}
    \caption{Comparison of generated results in 0-th task of MNIST-5T.}
    \label{fig:comparison_mnist}
\end{figure}


\begin{figure*}[h]
    \centering
    \begin{subfigure}{0.48\textwidth}
        \centering
        \includegraphics[width=\textwidth]{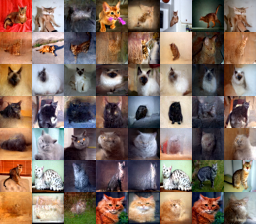}
        \caption{The Original Results.}
    \end{subfigure}
    \hfill
    \begin{subfigure}{0.48\textwidth}
        \centering
        \includegraphics[width=\textwidth]{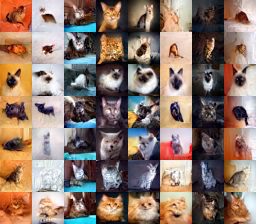}
        \caption{Our CCD Results.}
    \end{subfigure}
    \hfill
    \begin{subfigure}{0.48\textwidth}
        \centering
        \includegraphics[width=\textwidth]{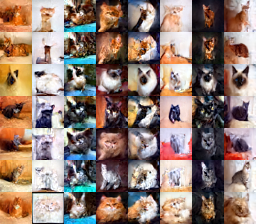}
        \caption{ER Results.}
    \end{subfigure}
    \hfill
    \begin{subfigure}{0.48\textwidth}
        \centering
        \includegraphics[width=\textwidth]{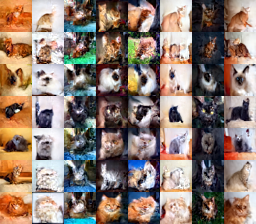}
        \caption{A-GEM Results.}
    \end{subfigure}
    \caption{Comparison of generated results in the 0-th task of OxfordPets-5T.}
    \label{fig:comparison_pets}
\end{figure*}

\begin{figure*}[h]
    \centering
    \begin{subfigure}{0.48\textwidth}
        \centering
        \includegraphics[width=\textwidth]{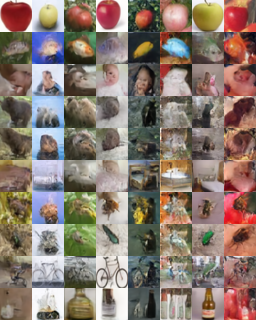}
        \caption{The Original Results.}
    \end{subfigure}
    \hfill
    \begin{subfigure}{0.48\textwidth}
        \centering
        \includegraphics[width=\textwidth]{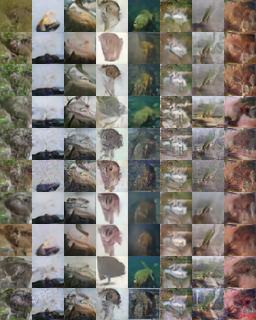}
        \caption{Our CCD Results.}
    \end{subfigure}
    \hfill
    \begin{subfigure}{0.48\textwidth}
        \centering
        \includegraphics[width=\textwidth]{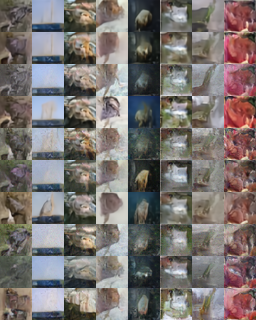}
        \caption{ER Results.}
    \end{subfigure}
    \hfill
    \begin{subfigure}{0.48\textwidth}
        \centering
        \includegraphics[width=\textwidth]{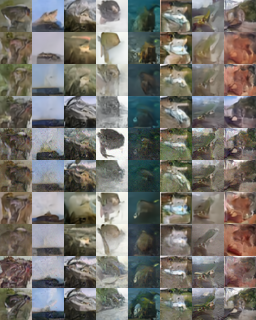}
        \caption{A-GEM Results.}
    \end{subfigure}
    \caption{Comparison of generated results in the 0-th task of CIFAR100-10T.}
    \label{fig:comparison_cifar}
\end{figure*}

\begin{figure*}[h]
    \centering
    \begin{subfigure}{0.48\textwidth}
        \centering
        \includegraphics[width=\textwidth]{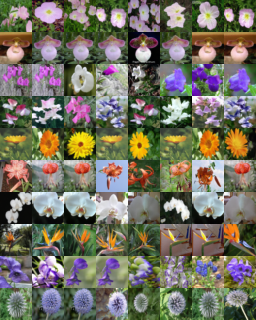}
        \caption{The Original Results.}
    \end{subfigure}
    \hfill
    \begin{subfigure}{0.48\textwidth}
        \centering
        \includegraphics[width=\textwidth]{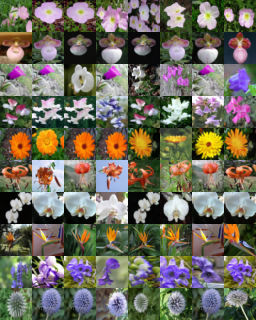}
        \caption{Our CCD Results.}
    \end{subfigure}
    \hfill
    \begin{subfigure}{0.48\textwidth}
        \centering
        \includegraphics[width=\textwidth]{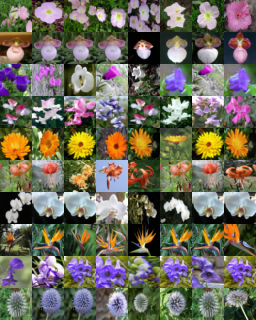}
        \caption{ER Results.}
    \end{subfigure}
    \hfill
    \begin{subfigure}{0.48\textwidth}
        \centering
        \includegraphics[width=\textwidth]{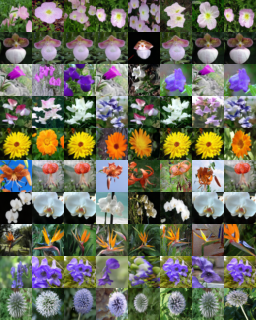}
        \caption{A-GEM Results.}
    \end{subfigure}
    \caption{Comparison of generated results in the 0-th task of Flowers102-10T.}
    \label{fig:comparison_flo}
\end{figure*}

\begin{figure*}[h]
    \centering
    \begin{subfigure}{0.24\textwidth}
        \centering
        \includegraphics[width=\textwidth]{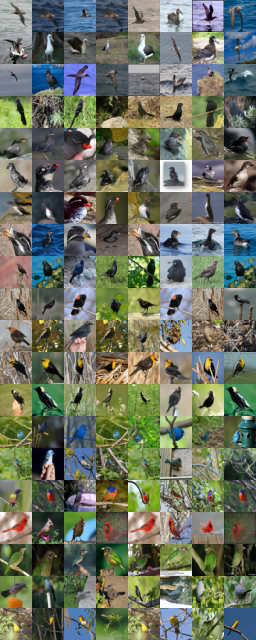}
        \caption{The Original Results.}
    \end{subfigure}
    \hfill
    \begin{subfigure}{0.24\textwidth}
        \centering
        \includegraphics[width=\textwidth]{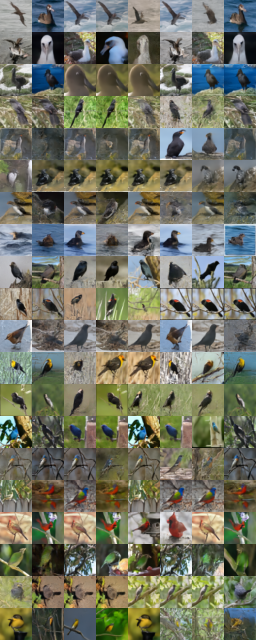}
        \caption{Our CCD Results.}
    \end{subfigure}
    \hfill
    \begin{subfigure}{0.24\textwidth}
        \centering
        \includegraphics[width=\textwidth]{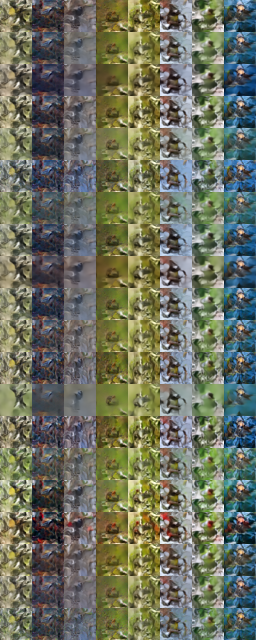}
        \caption{ER Results.}
    \end{subfigure}
    \hfill
    \begin{subfigure}{0.24\textwidth}
        \centering
        \includegraphics[width=\textwidth]{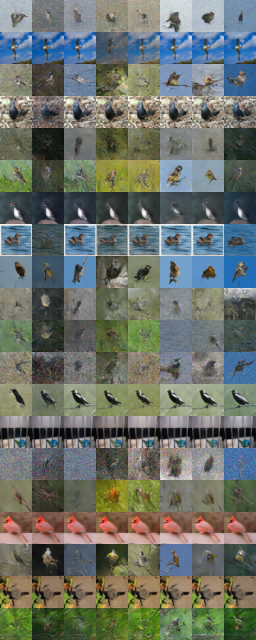}
        \caption{A-GEM Results.}
    \end{subfigure}
    \caption{Comparison of generated results in the 0-th task of CUB200-10T.}
    \label{fig:comparison_cub}
\end{figure*}

We conduct comprehensive evaluations across five continual generation benchmarks (MNIST-5T, OxfordPets-5T, CIFAR100-10T, Flowers102-10T, and CUB200-10T) to assess the effectiveness of our approach in retaining generative knowledge across tasks.
On MNIST-5T (Figure.~\ref{fig:comparison_mnist}), we observe that standard buffer-based baselines such as ER and A-GEM suffer from severe forgetting: they completely lose the ability to generate digits from the first task, including digits 0 and 1. In contrast, our method successfully reconstructs digit 0, evidencing improved knowledge retention. Nonetheless, the failure to accurately reproduce digit 1 suggests that GCF still persists, highlighting the need for more principled strategies for generative memory consolidation.

On the OxfordPets-5T dataset (Figure~\ref{fig:comparison_pets}), our method demonstrates clear improvements over both ER and A-GEM. The samples produced by ER and A-GEM suffer from significant distortions and blurring, particularly evident in columns 2–5, where the cats' faces often appear grotesquely warped and nearly unrecognizable. In contrast, our approach markedly reduces these artifacts, yielding substantially more realistic reconstructions. However, some residual imperfections in fine-grained details remain, suggesting that there is still considerable room for further enhancement.

On the more complex CIFAR100-10T dataset (Figure.~\ref{fig:comparison_cifar}), all compared methods, including ours, fail to retain generative knowledge from the initial task. This failure can be attributed to the minimal overlap in semantic content across tasks, making cross-task knowledge retention challenging. These results underscore a key limitation of current approaches, including ours: the reliance on shared structural information for knowledge retention. In scenarios where such structure is absent, task interference remains severe. This raises an important open question, how can we effectively preserve and transfer independent, task-specific knowledge without impeding the acquisition of new information?

In the Flowers102 benchmark (Figure.~\ref{fig:comparison_flo}), where the dataset size is comparable to the buffer capacity, all methods achieve moderate generative performance. However, qualitative differences are evident. Our model consistently generates samples with higher visual fidelity and stronger alignment to real data. For instance, it successfully captures rare instances, such as white flowers in the third category, that A-GEM entirely fails to reproduce. Moreover, in categories prone to error (e.g., the seventh category), our model avoids semantic drift and maintains accurate class representation, suggesting a stronger capacity for handling underrepresented classes.

Finally, on CUB200-10T, a fine-grained benchmark (Figure.~\ref{fig:comparison_cub}), our method clearly outperforms baselines in generative memory retention. It successfully reconstructs samples from the initial task, while ER and A-GEM fail to recover any meaningful representations. The alignment between fine-grained structure and our design principle of knowledge propagation yields consistently better generative fidelity. These findings not only validate our theoretical formulation but also demonstrate the practical advantage of our method in continual generation that demand nuanced representation learning.

In summary, our approach shows strong resilience to forgetting, particularly in tasks with shared visual structure or fine-grained semantics. However, its limitations in unstructured task regimes like CIFAR100-10T highlight the need for future work to better preserve task-specific knowledge in the absence of inter-task alignment.

\section{Future Improvements}
While our CCD framework demonstrates significant advances in CDG pipeline, several avenues for improvement emerge from our theoretical analysis and experimental findings:

\textbf{Adaptive Hyperparameter Tuning:} Our method relies on three key hyperparameters ($\kappa$, $\lambda$, $\eta$) whose optimal values exhibit dataset-dependent variation. Future work should investigate meta-learning approaches or automated hyperparameter optimization strategies to enhance cross-dataset robustness and reduce manual tuning overhead.

\textbf{Enhanced Buffer Construction:} Our experiments reveal that buffer quality often supersedes quantity, as evidenced by CIFAR100-10T where smaller buffers (512) outperform larger ones (2560 or 5120). Although our proposed HDB shows promise on coarse-grained datasets, it exhibits limitations on fine-grained tasks due to the non-discriminative nature of intermediate diffusion representations. Future research should focus on developing more sophisticated sample selection mechanisms that better capture the semantic diversity essential for effective continual generation.

\textbf{Minimal Cross-Task Overlap Scenarios:} A fundamental limitation of our approach lies in scenarios with minimal semantic overlap between tasks, as demonstrated by the challenging CIFAR100-10T results. Our consistency-based framework inherently relies on shared knowledge structures, making it less effective when such commonalities are absent. Developing strategies for knowledge retention and propagation under conditions of minimal cross-task alignment represents a critical research direction.

\textbf{Discriminative Representation Enhancement:} The effectiveness of memory-based methods in diffusion models is constrained by the non-discriminative nature of intermediate representations, which consist primarily of isotropic Gaussian noise. Future work should explore techniques to enhance the discriminative quality of diffusion latent spaces, potentially through architectural modifications that preserve semantic information throughout the denoising process.





\end{document}